\pdfoutput=1
\documentclass[11pt]{article}

\usepackage{acl}
\usepackage{times}
\usepackage{latexsym}

\usepackage[T1]{fontenc}

\usepackage[utf8]{inputenc}

\usepackage{microtype}

\usepackage{inconsolata}

\usepackage{graphicx}
\usepackage{algorithmic}
\usepackage{algorithm}
\usepackage{makecell}
\usepackage{multirow} 
\usepackage{mdframed}
\newmdenv[
  backgroundcolor=gray!10,
  linecolor=gray!10,
  leftmargin=10pt,
  rightmargin=10pt,
  skipabove=5pt,
  skipbelow=5pt
]{shadedquotation}
\usepackage{amsmath} 
\usepackage{adjustbox}

\usepackage{kotex}

\title{BridG MT: Enhancing LLMs' Machine Translation Capabilities\\with Sentence Bridging and Gradual MT}

\author{Seungwoo Choi, Gahyun Yoo, Jay-Yoon Lee*\\
        Seoul National University \\ \{rhdn520, padme0421, lee.jayyoon\}@snu.ac.kr}

\begin{document}
\maketitle
\begin{abstract}
Recent Large Language Models (LLMs) have demonstrated impressive translation performance without requiring fine-tuning on additional parallel corpora. However, they still face significant challenges in certain scenarios, particularly when translating low-resource languages. A common approach to address this issue is to provide external knowledge, such as few-shot examples, to assist LLMs in translating specific source sentences. However, this method is fundamentally limited by the quality or quantity of relevant sources, which cannot always be guaranteed. To reduce LLMs' reliance on external sources, we propose \textit{BridG MT}, a method that combines \textit{Sentence Bridging}, which generates a sequence of sentences as a bridge that gradually transition from easy-to-translate to more difficult, and \textit{Gradual MT}, which sequentially translates these sentences using earlier translations as few-shot examples for subsequent ones. Experiments conducted on four LLMs across seven languages demonstrate that our method effectively enhances translation performance, even outperforming translation methods that rely on a large number of few-shot examples.
\end{abstract}

\section{Introduction}
\label{sec:introduction}

\begin{figure}[t]
\begin{center}
\includegraphics[width= 0.95\columnwidth]{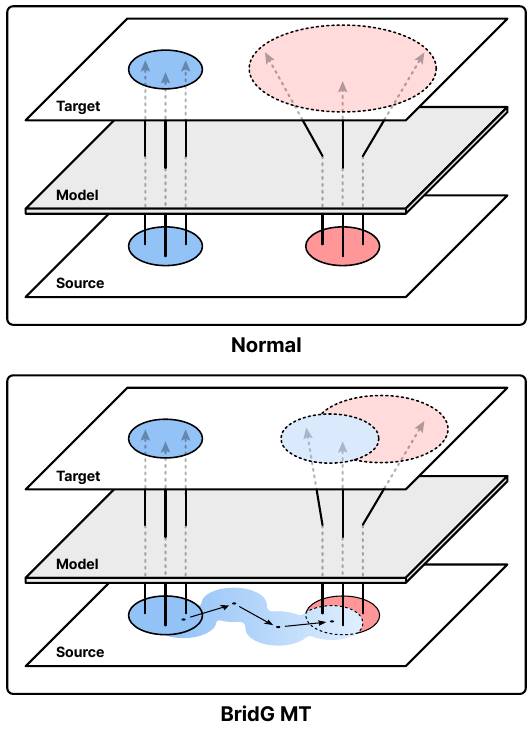}
\caption{Figurative illustration of BridG MT. Machine translation maps between source and target language spaces. BridG MT expands well-performing areas (blue) by leveraging neighboring examples, reaching poorly performing areas (red). Circle sizes indicate output noisiness.}
\label{fig:intuition}
\end{center}
\end{figure}

Recent Large Language Models (LLMs) have shown strong performance in translation tasks without the need for fine-tuning on specific parallel datasets. Previous studies have demonstrated that LLMs’ translation capabilities are reliable in most use cases, particularly when the source and target language are high-resource languages \citep{zhu_multilingual_2024, robinson_chatgpt_2023, jiao_is_2023}. However, because LLMs require training on large corpora, they still face challenges when translating low-resource languages that are not sufficiently represented in the training corpora.\citep{stap_chatgpt_2023, robinson_chatgpt_2023, enis_llm_2024}.

Previous research has attempted to address these challenges by leveraging the in-context-learning capabilities of large language models (LLMs), particularly through the use of external knowledge such as few-shot examples or dictionaries during inference. However, relevant examples are not always guaranteed to be available, and constructing such external knowledge sources can be costly. A potential solution is to reduce reliance on external sources altogether.

In this paper, we investigate whether LLMs' translation capabilities can be improved without relying on external knowledge. Our approach is grounded in a fundamental idea: machine translation can be understood as a mapping between two (sub)spaces representing the source and target languages. Within the source language space, some regions yield accurate translations, while others do not. The key question is whether we can expand the model's confident region by refeeding examples from it, ultimately extending into weaker regions and improving performance in those areas.

We propose \textit{BridG MT} to enhance machine translation by bridging regions where the LLM performs well with those where it struggles. BridG MT integrates two key techniques: \textit{Sentence Bridging} and \textit{Gradual MT}. Sentence Bridging is a prompting method that generates a sequence of sentences, progressively transitioning between them. Gradual MT iteratively translates a list of sentences, using the model's previous translations as few-shot examples for subsequent ones. Sentence Bridging helps bridge high- and low-performance regions, while Gradual MT gradually expands the model’s strong performance areas by leveraging in-context learning. These concepts are visually illustrated in Figure \ref{fig:intuition}.

We evaluate the effectiveness of BridG MT by applying it to four different LLMs: GPT-3.5, Mistral-Nemo-Instruct, Llama-3.1-70B-Instruct, and Llama-3.1-8B-Instruct. The experiments cover seven target languages: German (De), Chinese (Zh), Hindi (Hi), Korean (Ko), Swahili (Sw), Marathi (Mr), and Bengali (Bn). The results demonstrate that BridG MT significantly enhances the translation capabilities of LLMs, particularly in low-resource languages. To the best of our knowledge, we are the first to propose the concept of Sentence Bridging. 

\section{Related Work}

\subsection{Enhancing LLMs’ Translation Capabilities without Fine-tuning}
Modern LLMs demonstrate strong translation capabilities in high-resource languages but struggle with low-resource languages \citep{jiao_is_2023, stap_chatgpt_2023, zhu_multilingual_2024, enis_llm_2024}. Several studies have focused on enhancing LLMs’ translation performance without additional fine-tuning. A primary approach involves leveraging LLMs' ability to learn from demonstrations or descriptions through in-context learning \citep{brown_language_2020, wei_chain--thought_2022}. Researchers have explored methods such as selecting appropriate exemplars for few-shot learning and demonstrating linguistic knowledge \citep{agrawal_-context_2022, vilar_prompting_2023, zhang_hire_2024}. Beyond simply providing examples, some approaches conduct a prior analysis of the sentence to be translated, using LLMs to extract relevant information. These approaches then supplement the translation process with resources such as chains of multilingual dictionaries \citep{lu-etal-2024-chain}, or adopt human-like strategies by providing the LLM with keywords, topics, and generated demonstrations \citep{he-etal-2024-exploring-maps}. Other studies have instead focused on generating multiple translation candidates and selecting the optimal output through ranking mechanisms such Minimum Bayes Risk decoding \citep{fernandes-etal-2022-qa-decoding}.

\subsection{Self-Demonstration}
Manually generating appropriate exemplars for in-context learning can be resource-intensive. To address this challenge, previous studies have explored enabling models to generate their own few-shot examples for tasks such as classification \citep{lyu_z-icl_2023, kim_self-generated_2022} or other reasoning tasks \citep{zhang2023automatic, li_self-prompting_2024}. Our work is aligned with these efforts, as it also focuses on generating the model's own few-shot examples. However, these techniques have yet to be applied to translation, and BridG MT’s novel approach of intentionally bridging confident regions to current predictions remains unexplored.

\section{Methodology}
We introduce two major components of BridG MT—Sentence Bridging and Gradual MT—in Sections \ref{subsec:sentint} and \ref{subsec:gradMT}, respectively, and then describe how we combine them in Section \ref{subsec:overall}.

\begin{figure*}[ht]
\centering
\includegraphics[width=1\textwidth]{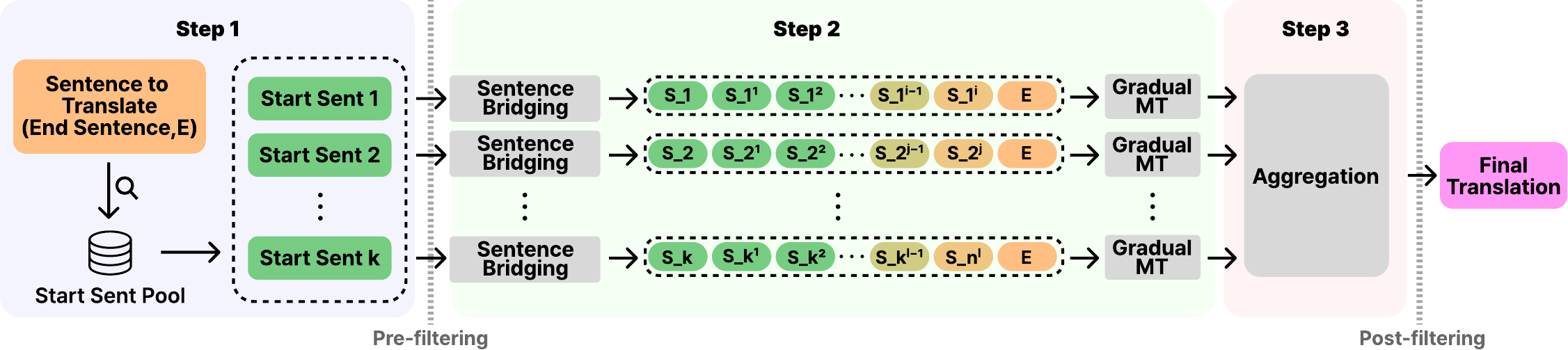} 
\caption{Illustration of the BridG MT algorithm. BridG MT integrates Sentence Bridging with Gradual MT. In Step 1, $k$ start sentences are selected from a predefined start-sentence pool. In Step 2, these start sentences are bridged to the end sentence, creating $k$ individual bridges. Each bridge is then processed through Gradual MT, generating translation results for every sentence along the path. In Step 3, the MT results from all bridges are aggregated into a single output translation. Optional pre- and post-filtering steps can be applied between Steps 1 and 2, and again after aggregation, to refine the sentences on which BridG MT is applied.}
\label{fig:bridg_mt}
\end{figure*}

\subsection{Sentence Bridging}
\label{subsec:sentint}
Sentence bridging is a prompting technique that asks the model to generate a list of sentences that gradually transition from a \textit{start sentence} to an \textit{end sentence}. The objective of this technique is to generate a list of sentences where each sentence is distinct, yet not excessively different from its adjacent sentences. We call these sentences the \textit{bridge}. In our experiments, we utilized three bridging examples from GPT-4\citep{openai_gpt-4_2024} as a few-shot to control the output format. See Appendices \ref{sec:intp_fewshots} and \ref{sec:intp_sample} for these examples and sample bridges. Following is the prompt that we use in our experiments:

\begin{quote}
I will give you two sentences. Can you gradually change the first sentence to make it exactly the same as the second sentence? Just give me the sentences and don't provide additional comments.\\
Sentence1: $\langle Sentence 1 \rangle$\\
Sentence2: $\langle Sentence 2 \rangle$    
\end{quote}

\subsection{Gradual MT}
\label{subsec:gradMT}
Gradual MT is a prompting technique that enables an LLM to leverage its previous translations as prompts. This approach sequentially processes a bridge, translating each source sentence while using the preceding translation results as few-shot examples for the current sentence. An illustration of the gradual MT algorithm is provided in Algorithm \ref{alg:grad_mt}. By applying gradual MT, we construct an expanding set of few-shot examples, which become increasingly useful for translating the final sentence that the model must process.

\begin{algorithm}[tb]
   \caption{Pseudo Code for Gradual MT}
   \label{alg:grad_mt}
\begin{algorithmic}
   \STATE {\bfseries Input:} Bridge $X=\{x_{1},x_{2},...,x_{n}\}$, $n = |X|$, Translation model $M$ 
   \STATE Set $fewshot= []$
   \FOR{$x_{i} \in X$}
   \STATE $\hat{y}_i = M(x_i, fewshot)$
   \STATE Append $\{x_i, \hat{y}_i\}$ to $fewshot$
   \ENDFOR
   \STATE{\bfseries Output:} $\hat{y}_{n}$
\end{algorithmic}
\end{algorithm}

\subsection{Overall Method}
\label{subsec:overall}
BridG MT combines sentence bridging and gradual MT. An illustration of the algorithm is shown in Figure \ref{fig:bridg_mt}.

\paragraph{Step 0: Start Sentence Pool Creation}
Before applying BridG MT, the \textit{start sentence pool} must first be created. This pool consists of sentences that the zero-shot LLM can translate with high accuracy. Since we are considering a scenario where there is no reliable gold translation, we utilize a reference-free quality estimation(QE) model such as CometKiwi\citep{rei_scaling_2023} to construct the start sentence pool. We refer to the source sentences in the start sentence pool as \textit{start sentences}.

\paragraph{Step 1: Start Sentence Selection}
BridG MT begins with selecting $k$ start sentences from the start sentence pool by calculating the similarity with the source sentence that the LLM is trying to translate, which we call an \textit{end sentence}. Measuring similarity between sentences can be done in various ways. In this paper, we utilized SBERT similarity\citep{reimers_sentence-bert_2019} as a primary metric. Details about start sentence selection strategy can be found in Section \ref{subsubsec:start_sentence_selection_strategy}.

\paragraph{Step 2: Sentence Bridging \& Gradual MT}
After selecting start sentences, sentence bridging is performed between each start sentence and the corresponding end sentence, creating $k$ separate \textit{bridges}. The bridges are then processed through the Gradual MT, generating translation results for each sentence.

\paragraph{Step 3: MT Results Aggregation}
When the number of start sentences is more than one, we proceed to aggregate the translation results from each bridge. This step is skipped if we choose to work with single start sentence in Step 1. There could be several methods for aggregating translations; in our approach, we input all the translation results into the LLM once again as few-shot examples to generate the final translation. A detailed ablation of this aggregation strategy can be found in Section \ref{subsubsec:Aggregation}. After aggregation, we obtain the final translation, which we refer to as the \textit{output}.

\paragraph{Pre- \& Post-filtering}
BridG MT can be applied to any sentence; however, it is often more effective to use it selectively for two reasons. First, determining which sentences will undergo BridG MT before its execution can significantly reduce computational costs. Second, even after applying BridG MT, it is better to discard the BridG MT result if its quality is worse than the zero-shot translation. We explored these two possibilities—referred to as Pre-filtering and Post-filtering—in Section \ref{subsubsec:filtering_strategy_ablation} and show that these methos can improve quality with smaller computation than full application of BridG MT.

\section{Experiment}
\subsection{Setup}
\paragraph{Models}
For \textbf{translation}, we use four different LLMs: GPT-3.5 (GPT-3.5-Turbo-0125)\footnote{\url{https://platform.openai.com/docs/models/gpt-3-5-turbo}}, Mistral-Nemo (Mistral-Nemo-Instruct-2407)\footnote{\url{https://huggingface.co/mistralai/Mistral-Nemo-Instruct-2407/}}, and two Llama models (Llama-3.1-70B-Instruct, Llama-3.1-8B-Instruct) \citep{dubey_llama_2024}. ChatGPT is accessed via OpenAI's API, while the other models ran on local GPU. For \textbf{sentence bridging}, we employ Qwen2-72B-Instruct \citep{yang_qwen2_2024} with 4-bit quantization as our primary bridging model. We also tested Llama-3.1-8B-Instruct and Llama-3.1-3B-Instruct to test the generalizability of the method in more cost-efficient settings. See Appendices \ref{sec:llm_translation_setting} and \ref{sec:llm_intp_setting} for settings for translation and bridging. For \textbf{pre- and post-filtering}, we utilize a reference-free QE model CometKiwi \citep{rei_scaling_2023} to avoid peeking at the gold translations. CometKiwi predicts a DA score, which rates translation quality on a scale from 0 to 100, normalized to a range of 0 to 1. Lastly, we used all-mpnet-base-v2\footnote{https://huggingface.co/sentence-transformers/all-mpnet-base-v2} for \textbf{SBERT sismilarity} calculation in retrieving start sentences and few-shot examples for baseline experiments. All experiments were conducted once.

\paragraph{Target Languages}
We fixed English as the source language. The target languages tested in the experiments are German (De), Chinese (Zh), Korean (Ko), Hindi (Hi), Swahili (Sw), Bengali (Bn), and Marathi (Mr). Based on \citet{joshi-etal-2020-state}'s 6 scale resource level, we classify German and Chinese as high resource, Korean and Hindi as mid resource, and the rest as low resource. For GPT 3.5, we experiment on every languages. For the Llama 3.1 models, we only experiment on German and Hindi, as they do not support other languages. Mistral-Nemo does not officially support Ko, Hi, Sw, Bn, and Mr, but we conduct experiments on those languages nevertheless, as it has some capability to generate them.

\paragraph{Dataset}
We use the FLORES-200 benchmark dataset \citep{nllb_team_no_2022} for validation and evaluation. The development split is used to construct the start sentence pool. For the test set, we sample 90\% of the data and reserve the remaining 10\% to determine the QE score threshold for pre-filtering. We also use NTREX-128 \citep{federmann-etal-2022-ntrex} to evaluate whether BridG MT remains effective when the distribution of the start sentence pool differs from that of the test set. The results for NTREX-128 are presented in Section \ref{sec:discussion}.

\paragraph{Start Sentence Pool Creation} 
The start sentence pool is created by translating source sentences from the dev split of the FLORES-200 dataset using a zero-shot approach. Each source sentence is translated five times and evaluated with CometKiwi-a reference-free model. The most frequently occurring translation is selected as the \textit{representative translation}. If no translation is repeated, the one with a score closest to the average is chosen. After selecting each representative translation, the top 100 translation pairs with the highest DA scores are selected.

\begin{table*}[ht]
\caption{xCOMET scores of BridG MT across different translation models and target languages. Sentence bridging was performed using Qwen2-72B-Instruct. Scores are multiplied by 100 for readability. For each MT model and language, the highest score is shown in bold and the second-highest is underlined. Cells are left blank when a specific model or method does not support the corresponding language.}
\label{tab:results_xCOMET}
\begin{center}
\resizebox{0.7\linewidth}{!}{
\begin{tabular}{ c | l | c  c | c  c | c  c  c }
 & & \multicolumn{2}{c|}{\textbf{High Resource}} & \multicolumn{2}{c|}{\textbf{Mid Resource}} & \multicolumn{3}{c}{\textbf{Low Resource}} \\
\textbf{MT Model} & \makecell[c]{\textbf{Method}} & \textbf{DE} & \textbf{ZH} & \textbf{HI} & \textbf{KO} & \textbf{SW} & \textbf{BN} & \textbf{MR} \\\hline
\multirow{5}{*}{GPT 3.5} & Zero-shot & 97.63 & 91.30 & 71.89 & 89.48 & 81.23 & 68.73 & 44.53 \\
 &MAPS&97.68&92.06&-&-&-&-&-\\
 &QA Decode&97.45&91.63&75.45&90.19&78.26&73.90&45.82\\\cline{2-9}
 & BridG {\tiny Post} & \textbf{98.04} & \textbf{92.63} & \textbf{77.90} & \textbf{92.57} & \textbf{83.65} & \textbf{75.41} & \textbf{53.11} \\
 & BridG {\tiny Pre \& Post} & \underline{97.86} & \underline{92.35} & \underline{77.63} & \underline{92.07} & \underline{83.34} & \underline{75.14} & \underline{51.51} \\\hline
\multirow{5}{*}{Llama 3.1 70B} & Zero-shot & 97.33 & - & 79.41 & - & - & - & - \\
 &MAPS&96.53&-&-&-&-&-&-\\
 &QA Decode&97.37&-&-&-&-&-&-\\\cline{2-9}
 & BridG {\tiny Post} & \textbf{97.93} & - & \textbf{84.45} & - & - & - & - \\
 & BridG {\tiny Pre \& Post} & \underline{97.73} & - & \underline{84.31} & - & - & - & - \\\hline
\multirow{5}{*}{Llama 3.1 8B} & Zero-shot & 94.99 & - & 69.93 & - & - & - & - \\
 &MAPS&79.10&-&-&-&-&-&-\\
 &QA Decode&88.08&-&\underline{77.67}&-&-&-&-\\\cline{2-9}
 & BridG {\tiny Post} & \textbf{97.25} & - & \textbf{78.30} & - & - & - & - \\
 & BridG {\tiny Pre \& Post} & \underline{96.99} & - &77.37& - & - & - & - \\\hline
\multirow{5}{*}{Mistral Nemo} & Zero-shot & 96.70 & 88.01 & 66.78 & 81.43 & 38.97 & 71.89 & 43.99 \\
 &MAPS&85.75&88.71&-&-&-&-&-\\
 &QA Decode&96.33&87.18&64.42&80.34&28.86&68.59&35.37\\\cline{2-9}
 & BridG {\tiny Post} & \textbf{97.70} & \textbf{90.99} & \textbf{73.95} & \textbf{89.45} & \textbf{45.67} & \textbf{78.35} & \textbf{57.24} \\
 & BridG {\tiny Pre \& Post} & \textbf{97.70} & \textbf{90.99} & \underline{73.82} & \underline{89.34} & \underline{45.18} & \underline{77.83} & \underline{54.55} \\\hline
 \multicolumn{2}{c|}{TowerInstruct} & 97.69 & 89.89 & - & 91.29 & - & - & - \\\hline
 \multicolumn{2}{c|}{NLLB} & 96.21 & 67.88 & 81.00 & 82.20 & 77.17 & 82.70 & 71.83 \\\hline
\end{tabular}}
\end{center}
\end{table*}

\begin{table*}[ht]
\caption{xCOMET scores of BridG MT across different translation models, sentence bridging models, and target languages. Scores are multiplied by 100 for readability, and the two highest scores for each MT model and language are highlighted in bold and underlined. For languages not supported by each model, the corresponding cells were left blank.}
\label{tab:results_diff_bridg_model_xCOMET}
\begin{center}
\resizebox{0.7\linewidth}{!}{
\begin{tabular}{ c | c | l | c | c | c | c | c | c | c }
\textbf{MT Model} & \makecell[c]{\textbf{Method}}  & \makecell[c]{\textbf{Bridging Model}} & \textbf{DE} & \textbf{ZH} & \textbf{HI} & \textbf{KO} & \textbf{SW} & \textbf{BN} & \textbf{MR} \\
\hline
\multirow{4}{*}{\textbf{GPT 3.5}} & \makecell[l]{\multirow{3}{*}{\textbf{BridG MT}}}& {\small Qwen2 72B Instruct} & \textbf{98.04} & \textbf{92.63} & \textbf{77.90} & \textbf{92.57} & \textbf{83.65} & \textbf{75.41} & \textbf{53.11} \\
& & {\small Llama 3.1 8B} & \underline{97.97} & \underline{92.38} & \underline{77.24} & \underline{91.85} & \underline{83.62} & 74.24 & \underline{52.27} \\
& & {\small Llama 3.2 3B} & 97.97 & 92.35 & 76.44 & 91.48 & 83.49 & \underline{74.35} & 52.13 \\\cline{2-10}
 & \makecell[l]{\textbf{Zero-shot}} & \makecell[c]{{\small N/A}} & 97.63 & 91.30 & 71.89 & 89.48 & 81.23 & 68.73 & 44.53 \\
\hline
\multirow{4}{*}{\textbf{Llama 3.1 70B}} & \makecell[l]{\multirow{3}{*}{\textbf{BridG MT}}} & {\small Qwen2 72B Instruct} & \textbf{97.93} & - & \textbf{84.45} & - & - & - & - \\
 & & {\small Llama 3.1 8B} & \underline{97.87} & - & \underline{83.97}& - & - & - & - \\
 & & {\small Llama 3.2 3B} & 97.85 &  & 83.65& - & - & - & - \\\cline{2-10}
 & \makecell[l]{\textbf{Zero-shot}} & \makecell[c]{{\small N/A}} & 97.33 & - & 79.41 & - & - & - & - \\
 \hline
 \multirow{4}{*}{\textbf{Llama 3.1 8B}} & \makecell[l]{\multirow{3}{*}{\textbf{BridG MT}}} & {\small Qwen2 72B Instruct} & \textbf{97.25} & - & \textbf{78.30}& - & - & - & - \\
 &&{\small Llama 3.1 8B} & 96.93 & - & \underline{77.83}& - & - & - & -  \\
 &&{\small Llama 3.2 3B} & \underline{97.14} &  & 77.65& - & - & - & -  \\\cline{2-10}
& \makecell[l]{\textbf{Zero-shot}} & \makecell[c]{{\small N/A}} & 94.99 & - & 69.93 & - & - & - & -  \\
\hline
\multirow{4}{*}{\textbf{Mistral-Nemo}} & \makecell[l]{\multirow{3}{*}{\textbf{BridG MT}}}&{\small Qwen2 72b Instruct} & \textbf{97.70} & \textbf{90.99} & \textbf{73.95} & \textbf{89.45} & \textbf{45.67} & \textbf{78.35} & \textbf{57.24} \\
 & &{\small Llama 3.1 8B} & \underline{97.52} & \underline{90.40} & \underline{72.34} & \underline{87.83} & \underline{43.33} & \underline{76.73} & \underline{52.93} \\
 & & {\small Llama 3.2 3B} & 97.42 & 90.04 & 71.62 & 86.55 & 42.30 & 75.83 & 51.66 \\\cline{2-10}
 & \makecell[l]{\textbf{Zero-shot}} & \makecell[c]{{\small N/A}} & 96.70 & 88.01 & 66.78 & 81.43 & 38.97 & 71.89 & 43.99 \\
\hline
\end{tabular}}
\end{center}
\end{table*}

\begin{figure*}[ht]
    \centering
    \includegraphics[width=1\linewidth]{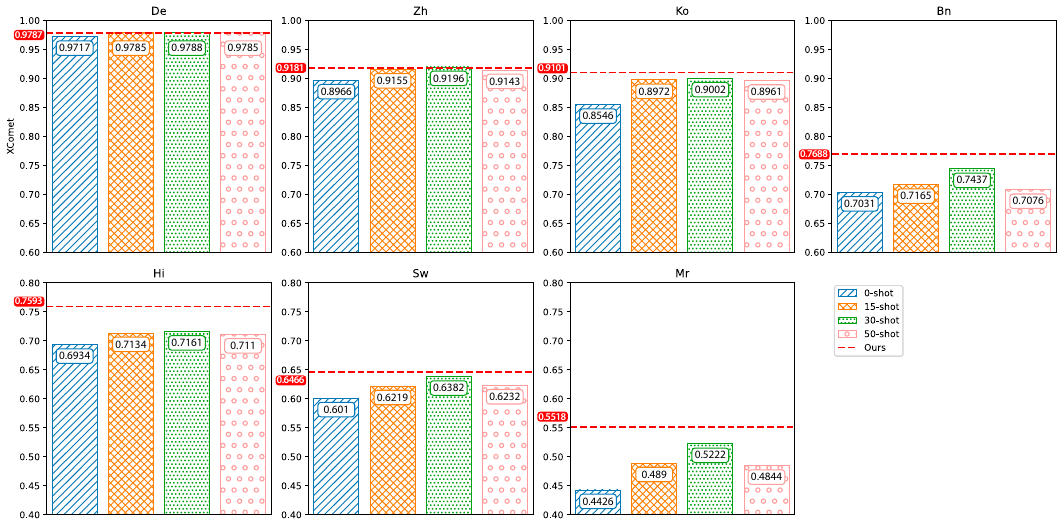}
    \caption{xCOMET scores for few-shot translations (bars) and BridG MT translations (horizontal line). Both BridG MT and few-shot scores represent the average xCOMET scores of GPT-3.5 and Mistral-Nemo.}
    \label{fig:fewshot_compare}
\end{figure*}

\label{sec:results}

\paragraph{Baselines} 
Since our focus is on scenarios where providing relevant few-shot examples is not feasible, we first compare BridG MT with zero-shot baselines. We then extend the comparison to conventional few-shot translation settings, where gold translations sampled from the start sentence pool are used as demonstrations (15, 30, and 50 examples). This enables us to evaluate how well BridG MT—designed specifically for settings without accessible few-shot examples—performs relative to standard few-shot approaches. We further compare BridG MT with related methods: MAPS \citep{he-etal-2024-exploring-maps}\footnote{https://github.com/zwhe99/MAPS-mt} and Quality Aware(QA) Decode \citep{fernandes-etal-2022-qa-decoding}\footnote{https://github.com/deep-spin/qaware-decode}. Among the target languages in our experiments, MAPS provides few-shot examples only for De and Zh in its published codebase; therefore, it was evaluated only on these languages. For QA Decode, we used MBR reranking, as it showed the best performance in their paper. Finally, to provide a broader context for comparison, we benchmark BridG MT against strong baseline models, including TowerInstruct 13B \citep{tower_llm_2024} and NLLB-200-3.3B\footnote{https://huggingface.co/facebook/nllb-200-3.3B}. Among the languages supported by TowerInstruct, only De, Zh, and Ko overlapped with our target languages; thus, experiments with TowerInstruct were conducted only on these languages.

\paragraph{Evaluation}
We employ four different QE models for evaluation; xCOMET, MetricX\citep{juraska_MetricX-23_2023}, CometKiwi, and BLEURT\citep{bleurt}. xCOMET, CometKiwi and BLEURT predict DA score, and MetricX predicts MQM score. MQM score assesses translation erorrs on a scale from 0 to 25, where lower scores indicate higher quality. Lastly, we report BLEU and ChrF.

\section{Results}
We first conducted a series of ablation experiments for En-Ko translation using GPT 3.5 to identify the optimal configuration, including start selection (Step 1) and result aggregation (Step 3). This optimal configuration was then applied in the main experiments across different translation models, sentence bridging models, and target languages. A detailed explanation of the ablation study is provided in Section \ref{subsec:ablation_study}. In this section, we present the xCOMET results of BridG MT across various LLMs and target languages. The full results are provided in Appendix \ref{sec:full_result_diff_metrics}.

\subsection{Comparing with Zeroshot Baselines}
As shown in Table \ref{tab:results_xCOMET}, BridG MT consistently improves the translation performance of LLMs, particularly for low-resource languages. For example, the performance gaps of GPT 3.5 between BridG MT and the baseline are 0.4, 1.3, 6.0, 3.1, 2.4, 6.7, and 8.6 for De, Zh, Hi, Ko, Sw, Bn, and Mr, respectively. This trend remains consistent across other evaluation metrics (Tables \ref{tab:results_full_cometkiwi}, \ref{tab:results_full_bleurt}, \ref{tab:results_full_metricx}, \ref{tab:results_full_bleu}, and \ref{tab:results_full_chrf}). BridG MT also outperforms alternative methods—MAPS and QA Decode. Notably, similar patterns are observed even when using QE models other than COMET and when varying the choice of bridging model. As shown in Table \ref{tab:results_diff_bridg_model_xCOMET}, using smaller bridging models leads to a slight decline in performance, but the overall improvement over the baseline remains substantial.

\subsection{Comparing with Baselines Using Labelled Data}
We further compared our results to a few-shot setup. To provide few-shot examples under conditions similar to our approach, we built a few-shot pool by selecting 100 sentences from the FLORES-200 dev set that the model translates well—similar to how we created the start sents pool. However, in this case, we employed xCOMET, a reference-based QE model. Figure \ref{fig:fewshot_compare} presents the average translation scores of GPT-3.5 and Mistral-Nemo. The results indicate that our method is comparable to, and often outperforms, few-shot translations. This demonstrates that even without high-quality gold translations, BridG MT can surpass translations generated by an LLM equipped with gold few-shot references. Additionally, we observe a performance drop when the model is given 50-shot examples, suggesting that BridG MT achieves performance levels that cannot be matched simply by increasing the number of few-shot examples. These findings remain consistent when using Llama models and evaluating with other metrics. See Appendix \ref{sec:full_result_diff_metrics} for detailed results.

\section{Ablation Study}
\label{subsec:ablation_study}
To find the optimal configuration for the BridG MT, we conducted an ablation study on En-Ko translation task using GPT-3.5 in terms of four different dimensions: Start sentence selection criteria, Number of start sentences, MT result aggregation strategy, and Filtering strategy. We utilized xCOMET with gold translations in the ablation study to construct the highest-quality start pool possible. Full results with all different combination of ablation settings can be found in Appendix \ref{sec:ablation_full_result}. 

\subsection{Start Sentence Selection Strategy}
\label{subsubsec:start_sentence_selection_strategy}

Based on the intuition that start sentences similar to the source sentence will be helpful, we employed three different metrics—SBERT similarity \citep{reimers_sentence-bert_2019}, Levenshtein distance \citep{levenshtein_binary_1966}, and tree edit distance \citep{zhang_simple_1989}—to calculate the similarity between sentences, and we combined them in three different ways. The first approach, \textit{Sort}, sorts sentences using multiple metrics with varying priorities. The second approach, \textit{Filter}, initially selects the top 10 sentences based on SBERT similarity and then sorts that selection using the other metrics. The third approach, \textit{Tops}, picks the top sentence based on the highest similarity scores from each metric. As shown in Table \ref{tab:sss}, the selection strategies that produced the highest scores differed across aggregation methods. We chose to sort by SBERT similarity and then by tree edit distance (Sort(S-T)), as this approach yielded the highest average scores.

\begin{table}[ht]
\caption{Averaged scores with different start selection strategies, start sentence numbers and aggregation strategies applied to EN-KO translation task. `Sort', `Filter', and `Tops' denote the start sentence selection methods. Each letter in parentheses represents a similarity metric, with the order indicating the priority of these metrics. `L', `T' and `S' stands for Levenshtein distance, Tree edit distance, and SBERT similarity, respectively. The highest average values for each axis are highlighted in bold. The highest values for each start sentence number and aggregation strategy are highlighted in underline.}
\label{tab:sss}
\begin{center}
\resizebox{1\linewidth}{!}{\begin{tabular}{c c c c c}  
\multirow{2}{*}{\textbf{Start Selection Strategy}}  & \multicolumn{3}{c}{\makecell{\textbf{Start Sents.Num \&}\\ \textbf{Aggregation Strategy}}}& \multirow{2}{*}{\textbf{Average}}\\\cline{2-4}
& \textbf{1 (n/a)} & \textbf{3 Poll} & \textbf{3 Prompt} & \\\hline
\multicolumn{1}{c|}{Filter(T-L)} & 91.29 & 91.22 & 91.72 & \multicolumn{1}{|c}{91.41} \\
\multicolumn{1}{c|}{Filter(L-T)} & \underline{91.67} & 91.23 & 91.59 & \multicolumn{1}{|c}{91.50} \\
\multicolumn{1}{c|}{Sort(L-S)} & 91.35 & 91.22 & 91.57 & \multicolumn{1}{|c}{91.38} \\
\multicolumn{1}{c|}{Sort(T-S)} & 90.95 & 91.09 & 91.65 & \multicolumn{1}{|c}{91.23} \\
\multicolumn{1}{c|}{Sort(L-T-S)} & 91.19 & 91.21 & 91.50 & \multicolumn{1}{|c}{91.30} \\
\multicolumn{1}{c|}{Sort(T-L-S)} & 91.19 & 91.00 & \underline{91.93} & \multicolumn{1}{|c}{91.37} \\
\multicolumn{1}{c|}{Sort(S-T)} & 91.50 & \underline{91.39} & 91.83 & \multicolumn{1}{|c}{\textbf{91.57}} \\
\multicolumn{1}{c|}{Tops} &  -& 91.23 & 91.69 & \multicolumn{1}{|c}{91.46} \\\hline
\multicolumn{1}{c|}{\textbf{Average}} & 91.31 & 91.20 & \textbf{91.69} & \multicolumn{1}{|c}{}\\\hline
\end{tabular}}
\end{center}
\end{table}

\begin{table*}[ht]
\caption{Average xCOMET scores and score changes of selected outputs when applying the optimal strategies for start selection and aggregation in En-Ko translation. `Score change' is calculated only for the adopted outputs. We also report the number of end sentences for which bridging and Gradual MT is executed (`\# Bridging') and the number of end sentences for which the BridG MT output is selected over zero-shot translation (`\# Selected Outputs'). The results show that `Pre \& Post' reduces the number of bridging by more than half while maintaining nearly the same translation performance.}
\label{tab:adoption}
\begin{center}
\resizebox{0.7\linewidth}{!}{
\begin{tabular}{rrrrr}  
\makecell[l]{\textbf{Filtering}\\\textbf{Strategy}} &\makecell[l]{\textbf{Avg. Score}} & \makecell[l]{\textbf{Avg. Score Change} \\\textbf{of Selected Outputs}}& \makecell[l]{\textbf{\# Bridging}} & \makecell[l]{\textbf{\# Selected Outputs}}\\\hline
Zeroshot & 89.48 & - & - & -\\
All & 91.22 & 1.74 & \makecell[r]{911 (100\%)} & \makecell[r]{911 (100\%)}\\
Pre & 91.50 & 4.49 & \makecell[r]{410 (45\%)} & \makecell[r]{410 (45\%)}\\
Post & \textbf{92.54} & 5.61 & \makecell[r]{911 (100\%)} & \makecell[r]{497 (54\%)}\\
Pre \& Post & 92.08 & \textbf{8.50} & \makecell[r]{410 (45\%)} & \makecell[r]{279 (31\%)}\\\hline
\end{tabular}}
\end{center}
\end{table*}

\subsection{Number of Start Sentences \& MT Aggregation Strategy}
\label{subsubsec:Aggregation}
When selecting the start sentences, we need to decide whether to use more than one start sentence. If we choose more than one, we must aggregate each translation result generated by Gradual MT. We investigated two distinct strategies for aggregating the results of Gradual MT. The first method, referred to as \textit{Polling}, selects the MT result with the highest number of duplicates, drawing inspiration from prior research on self-consistency \citep{wang2023selfconsistency}. If no duplicates are found, a result is selected randomly. The second method, \textit{Prompting}, involves feeding all Gradual MT results into the LLM as few-shot examples to generate the final MT output. As shown in Table \ref{tab:sss}, the prompting strategy outperforms polling by 0.49 points. Polling is often worse than using a single start sentence.

\subsection{Filtering Strategy}
\label{subsubsec:filtering_strategy_ablation}
We tested three strategies for filtering. The first strategy, `Pre-filtering', aims to minimize costs by evaluating zero-shot translation results with a QE model and applying BridG MT only when necessary (i.e., when the QE score falls below a certain threshold). The second strategy, `Post-filtering', prioritizes maximizing performance by applying BridG MT first and using its output only if the QE score exceeds that of the zero-shot translation. The third strategy, `Pre- \& Post-filtering' combines the first two: applying BridG MT when the zero-shot translation's QE score is below a threshold, and only if BridG MT’s score is higher. We employed CometKiwi, a reference-free QE model, to implement these strategies. To compare them, we analyzed xCOMET scores and score changes of selected outputs, applying the optimal strategies for start selection (Sort(S-T)) and aggregation (Prompting). As shown in Table \ref{tab:adoption}, all strategies improved overall performance, with `Post-filtering' achieving a notable gain of over 1 point in QE scores compared to zero-shot MT. The results also indicate that `Pre- \& Post-filtering' reduces bridging by more than half while maintaining comparable performance to `Post-filtering', offering an effective compromise between computational efficiency and translation quality. Results with every combination of ablation strategies are shown in Table \ref{tab:adoption_all}.

\subsection{Optimal Configuration}
Based on our ablation study, we found the optimal strategy to be selecting start sentences using SBERT similarity and tree edit distance (Sort(S-T)), using three start sentences, aggregating Gradual MT results by feeding them as few-shot examples for the final translation, and applying post-filtering. However, given the computational cost, using both pre- and post-filtering can also be a good compromise between efficiency and accuracy.

\section{Discussion}
\label{sec:discussion}
\subsection{Generalizability of Start Sentence Pool}
To test the robustness of the start sentence pool, we evaluated BridG MT on the NTREX-128 dataset, which has a different distribution from FLORES-200—the dataset used to construct the start pool. We used Llama-3.1-8B for sentence bridging. The results show that even when the same start pool is used to translate a different dataset, BridG MT still achieves a significant performance improvement. Detailed results are provided in Table \ref{tab:ntrex_result_xcomet}.

\begin{table*}[ht]
\caption{xCOMET results of BridG MT on the NTREX-128 dataset for various target languages.}
\label{tab:ntrex_result_xcomet}
\begin{center}
\resizebox{0.7\linewidth}{!}{
\begin{tabular}{c|c|c|c|c|c|c|c|c}
\textbf{MT Model} & \textbf{Setting} & \textbf{De} & \textbf{Ch} & \textbf{Hi} & \textbf{Ko} & \textbf{Sw} & \textbf{Be} & \textbf{Mr} \\
\hline
\multirow{2}{*}{GPT-3.5} & Zeroshot & 95.10 & 83.71 & 67.54 & 81.85 & 78.98 & 62.62 & 41.55 \\
 & BridG MT & \textbf{95.74} & \textbf{85.48} & \textbf{71.95} & \textbf{84.90} & \textbf{81.49} & \textbf{69.04} & \textbf{48.40} \\
\hline
\multirow{2}{*}{Llama-70B} & Zeroshot & 94.39 & - & 75.58 & - & - & - & - \\
 & BridG MT & \textbf{95.50} & - & \textbf{80.03} & - & - & - & - \\
\hline
\multirow{2}{*}{Llama-8B} & Zeroshot & 91.33 & - & 64.94 & - & - & - & - \\
 & BridG MT & \textbf{93.79} & - & \textbf{73.47} & - & - & - & - \\
\hline
\multirow{2}{*}{Mistral Nemo} & Zeroshot & 91.02 & 74.36 & 61.13 & 65.10 & 32.99 & 63.66 & 35.19 \\
 & BridG MT & \textbf{94.74} & \textbf{80.76} & \textbf{68.45} & \textbf{78.33} & \textbf{39.61} & \textbf{72.25} & \textbf{47.19} \\
\hline

\end{tabular}
}
\end{center}
\end{table*}

\subsection{Balancing Performance and Speed in BridG MT}
\label{subsec:cost_analysis} 
BridG MT is useful in that it can significantly enhance the translation performance of LLMs. However, the additional inference time introduced by sentence bridging and Gradual MT can be a drawback in scenarios where fast translation is required. In such cases, it may be more important to prioritize speed, even at the cost of a slight drop in performance. We analyzed the cost-performance trade-offs of three approaches to reduce cost: using a smaller bridging model, applying bridge sampling and pre-filtering. Assuming a setting where sentences are translated one at a time, we measured the time spent on each step of BridG MT and analyzed the inference time and performance across different configurations. A detailed breakdown of the inference time and cost analysis results can be found in Appendix \ref{sec:computational_cost}.

\paragraph{Smaller Model for Sentence Bridging}
Using smaller models can significantly reduce the time required for sentence bridging. As shown in Table \ref{tab:bridging_model_compare}, even with these smaller models, our approach achieves performance comparable to that of Qwen2-72B-Instruct while reducing the computation time by more than half.

\paragraph{Bridge Sampling}
Bridge sampling is a method to reduce the computational cost by applying Gradual MT only to the first(start), middle, and last(end) sentences along the interpolation path. We tested the bridge sampling using only one start sentence. Bridge sampling reduces the time for Gradual MT by 78\% while showing higher performance than baselines. Results are shown in Table \ref{tab:bridge_sampling_compare}.

\paragraph{Pre-Filtering}
Pre-filtering reduces the overall computational cost by applying BridG MT only to sentences that the model struggles to translate well. While pre-filtering introduces an additional cost for computing QE scores, this is outweighed by the overall reduction in computational cost. According to our statistics, pre-filtering reduced the overall computation time by 28\%. Results are shown in Table \ref{tab:filtering_comparison}.

\paragraph{Results from Cost-Efficient Configuration} 
We further tested the cost-efficient configuration, in which we use Llama-3.1-8B for both sentence bridging and translation, and applied bridge sampling. We also applied pre-filtering, as we had already confirmed its effectiveness in our ablation study (Section \ref{subsubsec:filtering_strategy_ablation}). The test result on Hindi shows that applying these strategies can reduce inference time by 63\%, while compromising performance by only 2.4 points.

\begin{table}[ht]
\caption{Comparison between the results of the main experiment (`Main') and a more efficient configuration (`Efficient'). The translation model used is Llama-3.1-8B, and the target language is Hindi.}
\label{tab:cost_efficient_setting}
\centering
\resizebox{1\columnwidth}{!}{
\begin{tabular}{c|l|c|c|c}
\multicolumn{2}{c|}{}&\textbf{Main} & \textbf{Efficient} & \textbf{Zeroshot} \\\hline
\multirow{4}{*}{\textbf{Components}} 
& Bridging Model  & Qwen2 72B & Llama 8B & - \\
& Pre-filtering    & X         & O        & - \\
& Path Sampling    & X         & O        & - \\
& Post-filtering   & O         & X        & - \\
\hline
\multicolumn{2}{c|}{\textbf{Avg. Inference Time}} & 46.59 & 17.19 & 1.74 \\\hline
\multicolumn{2}{c|}{\textbf{xCOMET}}         & 78.30 & 75.88 & 69.93 \\
\hline
\end{tabular}
}
\end{table}

\subsection{Sentence Bridging Analysis}
We conducted an analysis of the sentences within the sentence bridges. In particular, we examined whether the sentences within a bridge gradually become more similar to the end sentence on both the source and target sides as illustrated in Figure \ref{fig:intuition}. The analysis revealed that the embedding distance between the bridge sentences and the end sentence progressively decreases on the source side, and a similar trend is observed on the target side, where the embedding distance between the gold translation and the translations generated at each step of Gradual MT also decreases. A detailed analysis is provided in Appendix \ref{sec:analysis_sentence_bridging}.

\subsection{Quality of a Start Pool}
We analyzed the impact of the start pool quality using GPT-3.5 and Mistral-Nemo across four languages: Korean (Ko), Hindi (Hi), Swahili (Sw), and Marathi (Mr). For each language, we selected three distinct start pools from the FLORES-200 dev set, sorted by their translation xCOMET scores: \textit{High}, \textit{Mid}, and \textit{Low}. The \textit{High} start pool consists of the 100 sentences with the highest xCOMET scores. The \textit{Low} start pool includes the 100 sentences with the lowest scores, while the \textit{Mid} start pool is constructed using the 100 sentences located in the middle of the sorted dev set. We conducted BridG MT with a single start sentence, without applying any filtering method. The results show that the quality of the start pool had a degrading effect on the final translation when using Mistral-Nemo, whereas GPT-3.5 demonstrated robustness. Full results can be found in Appendix \ref{sec:startpool_qual_result}.

\section{Conclusion}
In this paper, we proposed BridG MT, a novel method to enhance the machine translation capabilities of various LLMs. BridG MT leverages sentence bridging and gradual MT to guide models, eliciting stronger translation performance. Experimental results across various models and languages demonstrate that our approach consistently improves translation quality, particularly in low-resource languages. Our approach is practical in that it does not require extra training and does not conflict with previous methods that utilize other kinds of prompting techniques.

\section*{Limitations}
BridG MT has a limitation in that it introduces computational overhead due to Sentence Bridging and the recursive nature of Gradual MT. While we have tested various methods to minimize the cost of BridG MT, such as pre-filtering and bridge sampling, future work could focus on further optimizing computational efficiency. Additionally, since the model used for sentence bridging did not perform well in languages other than English, we had to limit our study to cases where English was the source language. Exploring better prompting techniques to interpolate non-English sentences presents a promising direction for future research.

\section*{Acknowledgments}

This work was supported in part by the National Research Foundation of Korea (NRF) grant (RS-2023-00280883, RS-2023-00222663); by the National Super computing Center with super computing resources including technical support (KSC-2023-CRE-0176, KSC-2024-CRE-0065); by the Korea Institute of Science and Technology Information (KISTI) in 2025 (No.(KISTI) K25L1M1C1), aimed at developing KONI (KISTI Open Neural Intelligence), a large language model specialized in science and technology; and by the Institute of Information \& communications Technology Planning \& Evaluation (IITP) grant funded by the Korea government(MSIT) (RS-2025-02263754); by the Korea Medical Device Development Fund grant funded by the Korea government (the Ministry of Science and ICT, the Ministry of Trade, Industry and Energy, the Ministry of Health \& Welfare, the Ministry of Food and Drug Safety, 2025의약안003); partially supported by New Faculty Startup Fund from Seoul National University. We thank Deun Lee of \textit{Pixel Bridge} for her professional assistance in creating the figures used in this paper.

\bibliography{custom}

\appendix

\section{Few-shot Examples for Sentence Bridging}
\label{sec:intp_fewshots}
In this study, we gave three examples to the bridging model. We arbitrarily picked start and end sentences from the New York Times news articles. 

\subsection{Example 1}
\begin{quote}
    
\textbf{Sentence 1:} Surprisingly often, animals show up uninvited at sporting events. Sometimes, it gets a little weird.\\
\textbf{Sentence 2:} D. Cohen tried to parry attacks on his honesty as Todd Blanche, Donald J. Trump’s lawyer, tried to destroy his credibility with jurors.
\\
\textbf{Bridge}
\begin{enumerate}
    \item Surprisingly often, animals show up uninvited at sporting events. Sometimes, it gets a little weird.
    \item Often, unexpected events occur during live events, and sometimes these can get quite weird.
    \item At public events, such as sports or trials, unexpected guests often cause disturbances, leading to weird situations.
    \item In public spectacles, like sports or courtrooms, unexpected participants can disrupt the normal proceedings in strange ways.
    \item During public hearings, like in court, surprising accusations and defenses can emerge, often causing odd disruptions.
    \item In courtroom battles, unexpected arguments and witnesses can often turn the proceedings weird as each side tries to undermine the other.
    \item In legal battles, lawyers frequently confront each other with surprising tactics to challenge credibility, which can make the proceedings seem strange.
    \item Michael D. Cohen, during his trial, encountered surprising tactics by Todd Blanche, Donald J. Trump’s lawyer, aiming to disrupt his credibility in unusual ways.
    \item Michael D. Cohen tried to parry attacks on his honesty as Todd Blanche, Donald J. Trump’s lawyer, tried to destroy his credibility with jurors.
\end{enumerate}
\end{quote}

\subsection{Example 2}
\begin{quote}
    
\textbf{Sentence 1:} Georgia’s Parliament overrode a presidential veto to give final approval on Tuesday evening to a contentious bill that has plunged the country into a political crisis and threatened to derail the pro-Western aspirations of many Georgians in favor of closer ties with Russia. \\
\textbf{Sentence 2:} At that time, the buildings were abandoned and inaccessible to the public, and the famous ferry that shuttled immigrants from the island to Manhattan sank in a storm in 1968.\\
\textbf{Bridge}\\
\begin{enumerate}
\item Georgia’s Parliament overrode a presidential veto to give final approval on Tuesday evening to a contentious bill that has plunged the country into a political crisis and threatened to derail the pro-Western aspirations of many Georgians in favor of closer ties with Russia. 
\item The decision made by Georgia’s Parliament on Tuesday evening, to override a presidential veto and approve a controversial bill, has sparked a political crisis in the country and raised concerns about the impact on the country's pro-Western stance and relations with Russia.
\item The controversial bill approved by Georgia’s Parliament on Tuesday, despite a presidential veto, has caused a political crisis and stirred debates about the country’s Western alignment and its relation with Russia.
\item The unilateral move by Georgia’s Parliament on Tuesday to push through a divisive bill despite opposition from the president has led to a political crisis and raised concerns about the nation's Western affiliations and ties with Russia.
\item The recent decision by Georgia’s Parliament to defy the presidential veto and pass a controversial bill has created a political turmoil, questioning the country’s Western orientation and relationship with Russia.
\item Georgia’s Parliament's decision to override the presidential veto and approve the contentious bill has caused a political crisis in the country, prompting discussions on the impact on its Western ties and relations with Russia.
\item Georgia’s Parliament’s act of rejecting the presidential veto and endorsing the controversial bill on Tuesday has resulted in a political crisis, with implications for the country's pro-Western stance and connections with Russia.
\item Georgia’s Parliament moved forward with finalizing a contentious bill on Tuesday, overriding the presidential veto, which has triggered a political crisis and raised concerns about the nation's Western aspirations and relationship with Russia.
\item At that time, the buildings were abandoned and inaccessible to the public, and the famous ferry that shuttled immigrants from the island to Manhattan sank in a storm in 1968.
\end{enumerate}
    
\end{quote}

\subsection{Example 3}
\begin{quote}
\textbf{Sentence 1:} The cake has become one of Jamaica’s most popular desserts, often found in various stores.\\\\
\textbf{Sentence 2:} Some plans might cover only generic versions of certain contraceptives, but patients are still entitled to coverage of a specific product that their providers deem medically necessary.\\\\
\textbf{Bridge}
\begin{enumerate}
\item The cake has become one of Jamaica’s most popular desserts, often found in various stores.
\item Some plans might only cover basic versions of popular desserts like the cake, while others are more comprehensive.
\item In many cases, plans might cover only generic versions of popular products, but specific varieties might still be available.
\item Some plans might cover only generic versions of certain products, with exceptions allowed for specific needs.
\item Some plans might cover only generic versions of certain healthcare products, but specific options are still accessible if necessary.
\item Some plans might cover only generic versions of certain contraceptives, but patients can still get specific types if deemed necessary.
\item Some plans might cover only generic versions of certain contraceptives, but patients are still entitled to coverage of a specific product that their providers deem medically necessary.
\end{enumerate}
    
\end{quote}

\section{Sample Sentence Bridge}
\label{sec:intp_sample}
Below is a sample of the bridge generated by Qwen2-72b-Intstruct during the experiment.

\subsection{Sample 1}

\begin{quote}
    
\textbf{Prompt}
\\
Sentence1: I am happy that there are people willing to support me.\\
Sentence2: ``We now have 4-month-old mice that are non-diabetic that used to be diabetic," he added.\\\\
\textbf{Output}
\begin{enumerate}
    \item I am happy that there are people willing to support me.
    \item It's good to know that support is available when needed.
    \item It's reassuring to see support being provided in various situations.
    \item He mentioned seeing positive changes in different scenarios.
    \item ``We have seen positive transformations in our subjects," he added.
    \item ``We now have improved conditions in our test subjects," he added.
    \item ``We now have 4-month-old mice that are non-diabetic that used to be diabetic," he added.
\end{enumerate}
\end{quote}

\subsection{Sample 2}
\begin{quote}
\textbf{Prompt}
\\Sentence1: The death toll is at least 15, a figure which is expected to rise.\\
Sentence2: The moisture on your hands will react with the outer layers, which will feel funny and form a sort of shell.\\\\
\textbf{Output}
\begin{enumerate}
    \item The death toll is at least 15, a figure which is expected to rise.
    \item The number affected is at least 15, an amount which could increase due to underlying factors.
    \item An initial count shows at least 15 impacted, with this figure potentially growing as outer layers are peeled back.
    \item A preliminary assessment reveals at least 15 affected individuals, with the outer layers revealing more as they are examined.
    \item The moisture on your hands will interact with the outer layers, which may reveal more than initially thought.
    \item The moisture on your hands will react with the outer layers, which will feel peculiar and begin to peel away.
    \item The moisture on your hands will react with the outer layers, which will feel funny and form a sort of shell.
\end{enumerate}
\end{quote}

\section{Computational Cost}
\label{sec:computational_cost}

Tables \ref{tab:breakdown_cost}, \ref{tab:bridging_model_compare}, \ref{tab:bridge_sampling_compare} show the breakdown of the computational cost of BridG MT and the costs of possible scenarios.

\begin{table*}[ht]
\caption{Breakdown of inference time per component. The statistics are calculated from the FLORES test set into Hindi using Llama-3.1-8b.}
\label{tab:breakdown_cost}
\centering
\resizebox{0.7\textwidth}{!}{
\begin{tabular}{l l}
\hline
\makecell[c]{\textbf{Step}} & \makecell[c]{\textbf{Average Computation Time per Sentence}} \\
\hline
\textbf{Start Sentence Selection (FAISS)} & 0.09s (for whole dataset) \\
\textbf{Pre-Filtering / Post-Filtering (batch size = 8)} & 6.22s \\
\multicolumn{2}{l}{\textbf{Sentence Bridging}} \\
\quad w/ Llama-3.2-3B-Instruct & 6.03s \\
\quad w/ LlaMa-3.1-8B-Instruct & 12.94s \\
\quad w/ Qwen2-72b-Instruct & 26.17s \\
\multicolumn{2}{l}{\textbf{Translation (Llama-3.1-8B-Instruct)}} \\
\quad Zero-shot & 1.74s \\
\quad Gradual MT & 10.28s \\
\quad Gradual MT (w/ bridge sampling) & 2.29s \\
\quad 3-shot (for the final output) & 2.18s \\
\multicolumn{2}{l}{\textbf{Baselines}} \\
\quad 0 shot & 1.74s \\
\quad 30 shot & 2.33s \\
\quad 50 shot & 2.65s \\
\hline
\end{tabular}}
\end{table*}

\begin{table*}[ht]
\caption{Comparison of sentence bridging models in terms of inference time and average xCOMET score across different MT models and languages.}
\label{tab:bridging_model_compare}
\begin{center}
\resizebox{0.7\linewidth}{!}{
\begin{tabular}{lccc}
\makecell[c]{\textbf{Bridging Model}} & \textbf{Time for Sentence Bridging} & \textbf{Time for Overall Process} & \textbf{Average Score} \\\hline
Qwen2-72B-Instruct & 26.17s & 44.41s & 81.37 \\
Llama-3-1-8B-Instruct & 12.94s (49.45\%) & 31.18s (70.21\%) & 80.40 \\
Llama-3-2-3B-Instruct & 6.03s (23.04\%) & 24.27s (54.65\%) & 80.00 \\\hline
0-shot Baseline & 1.74s & 1.74s & 76.35 \\
50-shot Baseline & 2.65s & 2.65s & 78.71 \\
\hline
\end{tabular}
}
\end{center}
\end{table*}

\begin{table*}[ht]
    \caption{Comparison of using the full bridge and bridge sampling in terms of inference time and xCOMET score across different MT models and languages.}
\label{tab:bridge_sampling_compare}
\centering
\resizebox{0.8\linewidth}{!}{
\begin{tabular}{l c c c}
\makecell[c]{\textbf{Bridge Sampling}} & \textbf{Time for Gradual MT} & \textbf{Time for Overall Process} & \textbf{Average Score (xCOMET)} \\
\hline
Full Path {\small 3 start sentence} & 10.28s & 44.41s & 81.37 \\
Bridge Sampling {\small 1 start sentence} & 2.29s (22.28\%) & 36.42s (82.01\%) & 79.96 \\\hline
0-shot Baseline & 1.74s & -- & 76.35 \\
50-shot Baseline & 2.65s & -- & 78.71 \\
\hline
\end{tabular}}
\end{table*}

\begin{table*}[ht]
\caption{Comparison of filtering strategies in terms of time and xCOMET score across different MT models and languages.}
\label{tab:filtering_comparison}
\centering

\resizebox{0.5\linewidth}{!}{
\begin{tabular}{lcc}
\makecell[c]{\textbf{Filtering}} & \textbf{Time for Overall Process} & \textbf{Average Score (xComet)} \\
\hline
post-filtering & 46.59s & 81.37 \\
pre-filtering & 33.46s & 80.25 \\\hline
0-shot Baseline & 1.74s & 76.35 \\
50-shot Baseline & 2.65s & 78.71 \\
\hline
\end{tabular}}
\end{table*}

\section{Prompts and Settings for Translation}
\label{sec:llm_translation_setting}

\subsection{ChatGPT}
ChatGPT(gpt-3.5-turbo-0125) was used via API for translation with the same prompt from OpenAI's official documentation.\footnote{https://platform.openai.com/docs/examples} Temperature and top\_p were set to 0.3 and 1, respectively. The actual prompt is as follows:

\begin{quote}
\textbf{System:} You will be provided with a sentence in English, and your task is to translate it into $\langle$ Target  Language $\rangle$.

\textbf{User:} $\langle$ Sentence $\rangle$
\end{quote}

\subsection{Llama-3.1 70B \& 8B}
Llama-3.1 Instruct models were run on one A6000 GPU, using transformers library. 70B model were 4-bit quantized.Temperature and top\_p were set to 0.6 and 0.9, respectively. The actual prompt is as follows:
\begin{quote}
\textbf{System:} You will be provided with a sentence in English, and your task is to translate it into $\langle$ Target  Language $\rangle$.

\textbf{User:} $\langle$ Sentence $\rangle$
    
\end{quote}

\subsection{Mistral-Nemo-Instruct-2407}
Mistral-Nemo-Instruct-2407 was run on one A6000 GPU, using transformers library. Temperature and top\_p were set to 0.6 and 0.9, respectively. The actual prompt is as follows:
\begin{quote}
\textbf{User:} You will be provided with a sentence in English, and your task is to translate it into $\langle$ Target  Language $\rangle$.\\
Sentence: $\langle$ Sentence $\rangle$
\end{quote}

\section{Settings for Sentence Bridging}
\label{sec:llm_intp_setting}
Qwen2-72B-Instruct model was used for main experiments. It was run on one A6000 GPU with 4-bit quantization using transformers library. Temperature and top\_p were set to 0.6 and 0.9, respectively. 

Llama-3.1-8B-Instruct model was used for additional experiments. It was run on one A6000 GPU using transformers library. Temperature and top\_p were set to 0.6 and 0.9, respectively.

\section{Full Results with Different Metrics}
\label{sec:full_result_diff_metrics}
Table \ref{tab:results_full_xcomet}, \ref{tab:results_full_cometkiwi}, \ref{tab:results_full_bleurt}, \ref{tab:results_full_metricx}, \ref{tab:results_full_bleu}, \ref{tab:results_full_chrf} shows full results of our main experiment with different metrics; xCOMET, CometKiwi, MetricX, BLEURT, BLEU and ChrF.

\begin{table*}
\caption{Full xCOMET results across different sentence bridging models, translation models and target languages.}
\label{tab:results_full_xcomet}
\begin{center}
\resizebox{1\linewidth}{!}{
\begin{tabular}{ l | ccccccc | cc | cc | ccccccc }
 & \multicolumn{7}{c|}{\textbf{GPT-3.5-turbo-0125}}&\multicolumn{2}{c|}{\textbf{Llama-3.1-70B}} &\multicolumn{2}{c|}{\textbf{Llama-3.1-8B}} & \multicolumn{7}{c}{\textbf{Mistral-Nemo}}\\
\hline
\makecell[c]{\textbf{SETTING}} & \textbf{DE} & \textbf{ZH} & \textbf{HI} & \textbf{KO} & \textbf{SW} & \textbf{BN} & \textbf{MR} & \textbf{DE} & \textbf{HI} & \textbf{DE} & \textbf{HI} & \textbf{DE} & \textbf{ZH} & \textbf{HI} & \textbf{KO} & \textbf{SW} & \textbf{BN} & \textbf{MR} \\
\hline
\textbf{BASELINE 0 Shot} & 97.63 & 91.30 & 71.89 & 89.48 & 81.23 & 68.73 & 44.53 & 97.33 & 79.41 & 94.99 & 69.93 & 96.70 & 88.01 & 66.78 & 81.43 & 38.97 & 71.89 & 43.99 \\
\textbf{BASELINE 15 Shot} & 98.01 & 92.16 & 73.13 & 90.73 & 81.59 & 69.70 & 45.54 & 97.10 & 77.95 & 96.26 & 73.70 & 97.68 & 90.94 & 69.54 & 88.71 & 42.79 & 73.60 & 52.26 \\
\textbf{BASELINE 30 Shot} & 97.92 & 91.87 & 72.98 & 90.48 & 81.62 & 68.58 & 45.33 & 96.73 & 74.67 & 96.55 & 73.74 & 97.72 & 90.79 & 69.84 & 88.06 & 42.59 & 74.10 & 52.52 \\
\textbf{BASELINE 50 Shot} & 97.99 & 91.95 & 72.85 & 90.93 & 82.10 & 67.72 & 44.84 & 96.56 & 72.68 & 96.42 & 73.36 & 97.72 & 90.92 & 69.35 & 88.29 & 42.53 & 73.80 & 52.04 \\\hline
\textbf{BridG MT w/ Qwen2-72b} &  &  &  &  &  &  &  &  &  &  &  &  &  &  &  &  &  &  \\
\makecell[r]{Post Filtering} & 98.04 & 92.63 & 77.90 & 92.57 & 83.65 & 75.41 & 53.11 & 97.93 & 84.45 & 97.25 & 78.30 & 97.70 & 90.99 & 73.95 & 89.45 & 45.67 & 78.35 & 57.24 \\
\makecell[r]{Pre \& Post Filtering} & 97.86 & 92.35 & 77.63 & 92.07 & 83.34 & 75.14 & 51.51 & 97.73 & 84.31 & 96.99 & 77.37 & 97.70 & 90.99 & 73.82 & 89.34 & 45.18 & 77.83 & 54.55 \\
\textbf{BridG MT w/ Llama-3.1-8b} &  &  &  &  &  &  &  &  &  &  &  &  &  &  &  &  &  &  \\
\makecell[r]{Post Filtering} & 97.97 & 92.38 & 77.24 & 91.85 & 83.62 & 74.24 & 52.27 & 97.87 & 83.97 & 96.93 & 77.83 & 97.52 & 90.40 & 72.34 & 87.83 & 43.33 & 76.73 & 52.93 \\
\makecell[r]{Pre \& Post Filtering} & 97.95 & 91.54 & 76.96 & 91.42 & 83.52 & 74.06 & 50.86 & 97.73 & 83.90 & 96.82 & 73.09 & 97.41 & 90.40 & 71.19 & 87.42 & 43.09 & 76.23 & 51.29 \\
\textbf{BridG MT w/ Llama-3.2-3b} &  &  &  &  &  &  &  &  &  &  &  &  &  &  &  &  &  &  \\
\makecell[r]{Post Filtering} & 97.97 & 92.35 & 76.44 & 91.48 & 83.49 & 74.35 & 52.13 & 97.85 & 83.65 & 97.14 & 77.65 & 97.42 & 90.04 & 71.62 & 86.55 & 42.30 & 75.83 & 51.66 \\
\makecell[r]{Pre \& Post Filtering} & 97.92 & 92.29 & 76.39 & 91.21 & 82.74 & 74.10 & 52.04 & 97.61 & 83.28 & 96.86 & 76.84 & 97.42 & 90.04 & 71.46 & 86.33 & 42.14 & 75.75 & 51.05 \\
\hline
\textbf{QA Decode}& 97.45&91.63&75.45&90.19&78.26&73.90&45.82&97.37&82.32&88.08&77.67	&96.33&87.18&64.42&80.34&28.86&68.59&35.37\\
\textbf{MAPS} &97.68	&92.07&	-&	-	&-&	-	&-	&96.53	&-	&79.10	&-	&85.75&	88.71&	-	&-	&-&	-&	-\\\hline
\end{tabular}}
\end{center}
\end{table*}

\begin{table*}
\caption{Full CometKiwi results across different sentence bridging models, translation models and target languages.}
\label{tab:results_full_cometkiwi}
\begin{center}
\resizebox{1\linewidth}{!}{
\begin{tabular}{ l | ccccccc | cc | cc | ccccccc }
 & \multicolumn{7}{c|}{\textbf{GPT-3.5-turbo-0125}}&\multicolumn{2}{c|}{\textbf{Llama-3.1-70B}} &\multicolumn{2}{c|}{\textbf{Llama-3.1-8B}} & \multicolumn{7}{c}{\textbf{Mistral-Nemo}}\\
\hline
\makecell[c]{\textbf{SETTING}} & \textbf{DE} & \textbf{ZH} & \textbf{HI} & \textbf{KO} & \textbf{SW} & \textbf{BN} & \textbf{MR} & \textbf{DE} & \textbf{HI} & \textbf{DE} & \textbf{HI} & \textbf{DE} & \textbf{ZH} & \textbf{HI} & \textbf{KO} & \textbf{SW} & \textbf{BN} & \textbf{MR} \\
\hline
\textbf{BASELINE 0 Shot} & 86.12 & 85.84 & 69.12 & 87.73 & 83.24 & 67.93 & 56.23 & 84.97 & 74.55 & 80.06 & 66.71 & 83.32 & 82.90 & 59.72 & 78.41 & 42.72 & 62.03 & 52.53 \\
\textbf{BASELINE 15 Shot} & 86.71 & 86.71 & 69.72 & 88.85 & 82.99 & 69.65 & 58.37 & 84.11 & 72.94 & 81.64 & 69.58 & 85.46 & 85.73 & 66.77 & 87.48 & 47.51 & 72.81 & 63.73 \\
\textbf{BASELINE 30 Shot} & 86.44 & 86.74 & 69.51 & 88.86 & 83.12 & 68.39 & 57.71 & 83.22 & 70.19 & 82.17 & 69.40 & 85.38 & 85.74 & 67.00 & 86.88 & 48.08 & 73.00 & 63.35 \\
\textbf{BASELINE 50 Shot} & 86.61 & 86.81 & 69.30 & 89.18 & 83.42 & 67.98 & 57.32 & 82.70 & 68.28 & 81.95 & 69.05 & 85.30 & 86.04 & 66.78 & 87.08 & 47.61 & 72.83 & 63.09 \\\hline
\textbf{BridG MT w/ Qwen2-72b} &  &  &  &  &  &  &  &  &  &  &  &  &  &  &  &  &  &  \\
\makecell[r]{Post Filtering} & 87.42 & 87.64 & 74.30 & 90.88 & 86.09 & 75.34 & 66.01 & 87.25 & 78.04 & 85.40 & 74.00 & 86.63 & 86.89 & 70.96 & 89.51 & 54.71 & 77.65 & 68.16 \\
\makecell[r]{Pre \& Post Filtering} & 86.80 & 87.23 & 74.12 & 90.65 & 85.92 & 75.09 & 64.94 & 86.33 & 77.96 & 84.44 & 73.52 & 86.63 & 86.89 & 70.85 & 89.48 & 54.40 & 77.26 & 66.82 \\
\textbf{BridG MT w/ Llama-3.1-8b} &  &  &  &  &  &  &  &  &  &  &  &  &  &  &  &  &  &  \\
\makecell[r]{Post Filtering} & 87.28 & 87.42 & 73.83 & 90.36 & 85.76 & 74.30 & 64.86 & 87.16 & 77.91 & 84.86 & 73.25 & 86.48 & 86.39 & 69.42 & 88.17 & 51.82 & 75.79 & 64.70 \\
\makecell[r]{Pre \& Post Filtering} & 87.20 & 86.27 & 73.66 & 90.12 & 85.73 & 74.07 & 64.01 & 86.28 & 77.82 & 84.53 & 70.18 & 86.11 & 86.39 & 68.75 & 87.98 & 51.49 & 75.42 & 63.82 \\
\textbf{BridG MT w/ Llama-3.2-3b} &  &  &  &  &  &  &  &  &  &  &  &  &  &  &  &  &  &  \\
\makecell[r]{Post Filtering} & 87.13 & 87.38 & 73.24 & 90.29 & 85.58 & 74.42 & 64.74 & 87.19 & 77.94 & 85.31 & 73.30 & 86.20 & 85.97 & 68.62 & 87.63 & 50.19 & 74.83 & 6371 \\
\makecell[r]{Pre \& Post Filtering} & 86.91 & 87.26 & 73.21 & 90.08 & 84.95 & 74.22 & 64.67 & 85.98 & 77.64 & 84.41 & 72.84 & 86.20 & 85.97 & 68.54 & 87.47 & 50.00 & 7480 & 6349 \\
\hline
\textbf{QA Decode} &85.07&	85.84&	68.52	&88.16&	82.98	&72.29&	63.61	&84.85&	73.31&	82.52&	69.60&	81.93&	82.20	&55.94	&77.56	&39.83&	60.05&	50.49\\
\textbf{MAPS} &85.61	&86.39	&-	&-	&-&	-	&-&	83.48&	-	&71.19&	-	&71.06&	84.61&	-&	-&	-&	-	&-\\\hline
\end{tabular}}
\end{center}
\end{table*}

\begin{table*}
\caption{Full BLEURT results across different sentence bridging models, translation models and target languages.}
\label{tab:results_full_bleurt}
\begin{center}
\resizebox{1\linewidth}{!}{
\begin{tabular}{ l | ccccccc | cc | cc | ccccccc }
 & \multicolumn{7}{c|}{\textbf{GPT-3.5-turbo-0125}}&\multicolumn{2}{c|}{\textbf{Llama-3.1-70B}} &\multicolumn{2}{c|}{\textbf{Llama-3.1-8B}} & \multicolumn{7}{c}{\textbf{Mistral-Nemo}}\\
\hline
\makecell[c]{\textbf{SETTING}} & \textbf{DE} & \textbf{ZH} & \textbf{HI} & \textbf{KO} & \textbf{SW} & \textbf{BN} & \textbf{MR} & \textbf{DE} & \textbf{HI} & \textbf{DE} & \textbf{HI} & \textbf{DE} & \textbf{ZH} & \textbf{HI} & \textbf{KO} & \textbf{SW} & \textbf{BN} & \textbf{MR} \\
\hline
\textbf{BASELINE 0 Shot} & 78.58 & 73.70 & 68.31 & 68.77 & 75.80 & 67.95 & 68.03 & 77.57 & 71.34 & 74.69 & 66.96 & 76.27 & 69.87 & 60.95 & 59.13 & 51.51 & 62.29 & 64.20 \\
\textbf{BASELINE 15 Shot} & 79.27 & 74.16 & 68.49 & 69.41 & 75.84 & 68.46 & 69.29 & 77.12 & 70.61 & 75.69 & 68.54 & 77.95 & 72.05 & 66.52 & 66.94 & 54.71 & 70.14 & 70.75 \\
\textbf{BASELINE 30 Shot} & 79.06 & 74.19 & 68.75 & 69.41 & 76.01 & 67.87 & 69.00 & 76.35 & 69.32 & 75.99 & 68.62 & 78.07 & 72.30 & 66.64 & 66.56 & 55.00 & 69.79 & 70.91 \\
\textbf{BASELINE 50 Shot} & 79.10 & 74.20 & 68.74 & 69.54 & 76.13 & 67.82 & 68.56 & 76.10 & 68.68 & 75.80 & 68.41 & 77.89 & 72.40 & 66.46 & 66.49 & 55.19 & 70.15 & 70.46 \\\hline
\textbf{BridG MT w/ Qwen2-72b} &  &  &  &  &  &  &  &  &  &  &  &  &  &  &  &  &  &  \\
\makecell[r]{Post Filtering} & 79.11 & 74.11 & 69.91 & 70.28 & 76.51 & 70.85 & 71.22 & 78.56 & 72.39 & 76.98 & 69.20 & 77.91 & 71.38 & 66.85 & 67.17 & 56.50 & 71.71 & 71.40 \\
\makecell[r]{Pre \& Post Filtering} & 78.88 & 73.99 & 69.80 & 70.05 & 76.40 & 70.70 & 70.84 & 78.17 & 72.35 & 76.49 & 68.92 & 77.91 & 71.38 & 66.82 & 67.15 & 56.39 & 71.41 & 70.66 \\
\textbf{BridG MT w/ Llama-3.1-8b} &  &  &  &  &  &  &  &  &  &  &  &  &  &  &  &  &  &  \\
\makecell[r]{Post Filtering} & 79.13 & 74.19 & 69.75 & 69.99 & 76.31 & 70.54 & 70.65 & 78.61 & 72.28 & 76.75 & 68.88 & 77.67 & 71.04 & 65.94 & 66.00 & 55.16 & 70.71 & 70.04 \\
\makecell[r]{Pre \& Post Filtering} & 79.11 & 73.85 & 69.69 & 69.59 & 76.24 & 70.41 & 70.32 & 78.19 & 72.33 & 76.61 & 67.85 & 77.48 & 71.04 & 65.58 & 65.79 & 55.14 & 70.42 & 69.56 \\
\textbf{BridG MT w/ Llama-3.2-3b} &  &  &  &  &  &  &  &  &  &  &  &  &  &  &  &  &  &  \\
\makecell[r]{Post Filtering} & 78.94 & 74.16 & 69.51 & 69.62 & 76.31 & 70.10 & 70.61 & 78.55 & 72.30 & 76.93 & 68.84 & 77.65 & 70.94 & 65.59 & 65.03 & 54.61 & 70.02 & 69.45 \\
\makecell[r]{Pre \& Post Filtering} & 78.89 & 74.10 & 69.51 & 69.52 & 76.08 & 70.01 & 70.59 & 78.07 & 72.28 & 76.42 & 68.69 & 77.65 & 70.94 & 65.53 & 6492 & 54.56 & 70.00 & 69.33 \\
\hline
\textbf{QA Decode} &78.72	&74.07&69.83&69.45&76.60&71.63&72.04&78.41&72.67&76.78&70.00&76.11&69.81&60.85&59.18&51.42&61.93&64.32\\
\textbf{MAPS} & 79.08	&74.26	&-	&-&	-&	-	&-&	77.40&	-	&70.81&	-&	63.18&	71.69&	-&	-	&-&	-&	-\\\hline
\end{tabular}}
\end{center}
\end{table*}

\begin{table*}
\caption{Full MetricX results across different sentence bridging models, translation models and target languages.}
\label{tab:results_full_metricx}
\begin{center}
\resizebox{1\linewidth}{!}{
\begin{tabular}{ l | ccccccc | cc | cc | ccccccc }
 & \multicolumn{7}{c|}{\textbf{GPT-3.5-turbo-0125}}&\multicolumn{2}{c|}{\textbf{Llama-3.1-70B}} &\multicolumn{2}{c|}{\textbf{Llama-3.1-8B}} & \multicolumn{7}{c}{\textbf{Mistral-Nemo}}\\
\hline
\makecell[c]{\textbf{SETTING}} & \textbf{DE} & \textbf{ZH} & \textbf{HI} & \textbf{KO} & \textbf{SW} & \textbf{BN} & \textbf{MR} & \textbf{DE} & \textbf{HI} & \textbf{DE} & \textbf{HI} & \textbf{DE} & \textbf{ZH} & \textbf{HI} & \textbf{KO} & \textbf{SW} & \textbf{BN} & \textbf{MR} \\
\hline
\textbf{BASELINE 0 Shot} & 0.5864 & 1.1045 & 1.2189 & 0.5998 & 1.3240 & 2.2832 & 2.4860 & 0.6576 & 0.9091 & 1.0922 & 1.3274 & 0.6904 & 1.2751 & 1.6646 & 1.2064 & 7.2095 & 1.9082 & 2.6305 \\
\textbf{BASELINE 15 Shot} & 0.5428 & 1.0382 & 1.1840 & 0.5156 & 1.3116 & 2.0955 & 2.1551 & 0.7084 & 0.9696 & 0.7956 & 1.1012 & 0.5839 & 1.0667 & 1.2366 & 0.5462 & 6.3982 & 1.6427 & 1.6782 \\
\textbf{BASELINE 30 Shot} & 0.5477 & 1.0515 & 1.1629 & 0.5000 & 1.3188 & 2.1823 & 2.2366 & 0.7742 & 1.1978 & 0.7188 & 1.0863 & 0.5884 & 1.0718 & 1.2482 & 0.5842 & 6.3487 & 1.6230 & 1.8343 \\
\textbf{BASELINE 50 Shot} & 0.5532 & 1.0531 & 1.1319 & 0.5038 & 1.2999 & 2.2167 & 2.3578 & 0.7932 & 1.2621 & 0.7150 & 1.0899 & 0.5820 & 1.0725 & 1.2728 & 0.5568 & 6.5473 & 1.6864 & 1.8806 \\\hline
\textbf{BridG MT w/ Qwen2-72b} &  &  &  &  &  &  &  &  &  &  &  &  &  &  &  &  &  &  \\
\makecell[r]{Post Filtering} & 0.5432 & 1.0154 & 0.9179 & 0.4343 & 1.1661 & 1.5156 & 1.4508 & 0.5701 & 0.6854 & 0.6825 & 0.8481 & 0.5639 & 1.0808 & 1.0749 & 0.5559 & 5.3817 & 1.1181 & 1.2351 \\
\makecell[r]{Pre \& Post Filtering} & 0.5609 & 1.0370 & 0.9264 & 0.4460 & 1.1716 & 1.5285 & 1.5104 & 0.6058 & 0.6895 & 0.7356 & 0.8846 & 0.5639 & 1.0808 & 1.0809 & 0.5606 & 5.4107 & 1.1342 & 1.2881 \\
\textbf{BridG MT w/ Llama-3.1-8b} &  &  &  &  &  &  &  &  &  &  &  &  &  &  &  &  &  &  \\
\makecell[r]{Post Filtering} & 0.5494 & 1.0534 & 0.9157 & 0.4822 & 1.1922 & 1.6673 & 1.5956 & 0.5868 & 0.7184 & 0.7214 & 0.9050 & 0.5784 & 1.1299 & 1.2340 & 0.6131 & 5.8600 & 1.3074 & 1.5470 \\
\makecell[r]{Pre \& Post Filtering} & 0.5497 & 1.0896 & 0.9205 & 0.4948 & 1.1984 & 1.6793 & 1.6469 & 0.6022 & 0.7205 & 0.7467 & 1.0581 & 0.6008 & 1.1299 & 1.2725 & 0.6371 & 5.8815 & 1.3304 & 1.5953 \\
\textbf{BridG MT w/ Llama-3.2-3b} &  &  &  &  &  &  &  &  &  &  &  &  &  &  &  &  &  &  \\
\makecell[r]{Post Filtering} & 0.5471 & 1.0415 & 0.9824 & 0.4673 & 1.2055 & 1.6778 & 1.6031 & 0.5681 & 0.7210 & 0.6756 & 0.8896 & 0.6017 & 1.1462 & 1.2412 & 0.7434 & 6.1683 & 1.3213 & 1.6552 \\
\makecell[r]{Pre \& Post Filtering} & 0.5532 & 1.0404 & 0.9827 & 0.4759 & 1.2334 & 1.6827 & 1.6037 & 0.6146 & 0.7326 & 0.7168 & 0.9233 & 0.6017 & 1.1462 & 1.2521 & 0.7620 & 6.1826 & 1.3214 & 1.6684 \\
\hline
\textbf{QA Decode} &0.5764&1.0660	&0.9071	&0.4992&1.2059&	1.3878&	1.2131&	0.5948&	0.7066&	0.7348&	0.7422&	0.6983&	1.2705&	1.6804&	1.1841&	7.2072&	1.9488&	2.5831\\
\textbf{MAPS} &0.5516	&1.0451&	-&	-	&-	&-&	-	&0.6939&	-	&5.9713&	-&	3.6419&	1.1743&	-&	-	&-	&-	&-\\\hline
\end{tabular}}
\end{center}
\end{table*}

\begin{table*}
\caption{Full BLEU results across different sentence bridging models, translation models and target languages.}
\label{tab:results_full_bleu}
\begin{center}
\resizebox{1\linewidth}{!}{
\begin{tabular}{ l | ccccccc | cc | cc | ccccccc }
 & \multicolumn{7}{c|}{\textbf{GPT-3.5-turbo-0125}}&\multicolumn{2}{c|}{\textbf{Llama-3.1-70B}} &\multicolumn{2}{c|}{\textbf{Llama-3.1-8B}} & \multicolumn{7}{c}{\textbf{Mistral-Nemo}}\\
\hline
\makecell[c]{\textbf{SETTING}} & \textbf{DE} & \textbf{ZH} & \textbf{HI} & \textbf{KO} & \textbf{SW} & \textbf{BN} & \textbf{MR} & \textbf{DE} & \textbf{HI} & \textbf{DE} & \textbf{HI} & \textbf{DE} & \textbf{ZH} & \textbf{HI} & \textbf{KO} & \textbf{SW} & \textbf{BN} & \textbf{MR} \\
\hline
\textbf{BASELINE 0 Shot} & 40.45 & 45.58 & 23.06 & 27.86 & 32.93 & 9.99 & 5.94 & 38.71 & 29.09 & 30.81 & 21.54 & 35.70 & 38.83 & 17.46 & 20.17 & 12.13 & 8.35 & 5.31 \\
\textbf{BASELINE 15 Shot} & 40.86 & 46.02 & 22.98 & 28.81 & 33.80 & 9.67 & 7.54 & 37.33 & 25.54 & 32.13 & 22.14 & 36.89 & 40.65 & 20.88 & 25.08 & 14.40 & 11.03 & 7.30 \\
\textbf{BASELINE 30 Shot} & 40.86 & 46.00 & 23.08 & 29.08 & 33.57 & 9.10 & 6.94 & 36.13 & 23.29 & 32.75 & 22.36 & 36.96 & 41.03 & 21.05 & 25.45 & 14.44 & 11.13 & 7.45 \\
\textbf{BASELINE 50 Shot} & 40.96 & 45.98 & 23.51 & 28.90 & 33.95 & 9.75 & 6.08 & 35.39 & 22.84 & 32.61 & 22.37 & 36.72 & 41.14 & 21.05 & 25.28 & 13.75 & 11.16 & 7.69 \\\hline
\textbf{BridG MT w/ Qwen2-72b} &  &  &  &  &  &  &  &  &  &  &  &  &  &  &  &  &  &  \\
\makecell[r]{Post Filtering} & 40.87 & 46.35 & 23.72 & 29.01 & 33.30 & 11.06 & 8.00 & 38.42 & 28.08 & 32.43 & 22.12 & 35.79 & 37.76 & 18.53 & 23.96 & 14.09 & 10.06 & 7.19 \\
\makecell[r]{Pre \& Post Filtering} & 40.65 & 46.05 & 23.60 & 28.65 & 33.18 & 10.94 & 7.96 & 38.68 & 28.13 & 32.29 & 22.19 & 35.79 & 37.76 & 18.52 & 23.92 & 13.96 & 10.13 & 6.85 \\
\textbf{BridG MT w/ Llama-3.1-8b} &  &  &  &  &  &  &  &  &  &  &  &  &  &  &  &  &  &  \\
\makecell[r]{Post Filtering} & 40.86 & 46.27 & 23.97 & 28.87 & 33.57 & 10.93 & 7.94 & 39.11 & 28.11 & 32.23 & 22.09 & 35.65 & 37.77 & 18.71 & 23.57 & 13.39 & 10.11 & 6.80 \\
\makecell[r]{Pre \& Post Filtering} & 40.82 & 45.65 & 23.87 & 28.22 & 33.46 & 10.81 & 7.79 & 39.08 & 28.25 & 32.15 & 21.97 & 36.23 & 37.77 & 18.51 & 23.83 & 13.35 & 10.07 & 6.47 \\
\textbf{BridG MT w/ Llama-3.2-3b} &  &  &  &  &  &  &  &  &  &  &  &  &  &  &  &  &  &  \\
\makecell[r]{Post Filtering} & 40.79 & 46.18 & 23.53 & 28.66 & 33.30 & 10.86 & 8.42 & 38.95 & 28.36 & 32.37 & 22.09 & 36.08 & 37.49 & 18.82 & 22.89 & 12.96 & 9.87 & 6.07 \\
\makecell[r]{Pre \& Post Filtering} & 40.68 & 46.01 & 23.55 & 28.33 & 33.19 & 10.70 & 8.47 & 38.79 & 28.43 & 32.07 & 22.20 & 36.08 & 37.49 & 18.73 & 23.07 & 12.87 & 9.84 & 6.10 \\
\hline
\textbf{QA Decode} &41.07	&46.02	&23.65&	28.54	&33.59&	11.16	&8.57&	39.56	&30.06	&34.51	&23.66&	35.81	&38.67	&17.46&	20.39&	12.11	&7.98	&5.40\\
\textbf{MAPS} &40.83&	46.15&	-&	-&	-&	-	&-&	38.45	&-	&11.04&	-	&17.74	&40.63&	-&	-	&-	&-&	-\\\hline
\end{tabular}}
\end{center}
\end{table*}

\begin{table*}
\caption{Full ChrF results across different sentence bridging models, translation models and target languages.}
\label{tab:results_full_chrf}
\begin{center}
\resizebox{1\linewidth}{!}{
\begin{tabular}{ l | ccccccc | cc | cc | ccccccc }
 & \multicolumn{7}{c|}{\textbf{GPT-3.5-turbo-0125}}&\multicolumn{2}{c|}{\textbf{Llama-3.1-70B}} &\multicolumn{2}{c|}{\textbf{Llama-3.1-8B}} & \multicolumn{7}{c}{\textbf{Mistral-Nemo}}\\
\hline
\makecell[c]{\textbf{SETTING}} & \textbf{DE} & \textbf{ZH} & \textbf{HI} & \textbf{KO} & \textbf{SW} & \textbf{BN} & \textbf{MR} & \textbf{DE} & \textbf{HI} & \textbf{DE} & \textbf{HI} & \textbf{DE} & \textbf{ZH} & \textbf{HI} & \textbf{KO} & \textbf{SW} & \textbf{BN} & \textbf{MR} \\
\hline
\textbf{BASELINE 0 Shot} & 66.91 & 39.28 & 50.42 & 34.72 & 62.20 & 42.82 & 37.01 & 65.77 & 55.83 & 60.95 & 48.38 & 63.49 & 33.81 & 42.18 & 27.60 & 41.66 & 38.18 & 34.12 \\
\textbf{BASELINE 15 Shot} & 67.16 & 39.90 & 50.15 & 35.58 & 62.88 & 42.71 & 39.71 & 64.91 & 52.34 & 61.64 & 49.09 & 64.26 & 35.68 & 46.90 & 32.24 & 43.19 & 44.58 & 39.71 \\
\textbf{BASELINE 30 Shot} & 67.15 & 39.82 & 50.71 & 35.66 & 62.67 & 42.09 & 38.62 & 64.25 & 50.53 & 61.73 & 49.00 & 64.31 & 36.25 & 47.16 & 32.39 & 43.27 & 44.67 & 39.77 \\
\textbf{BASELINE 50 Shot} & 67.05 & 39.78 & 50.60 & 35.76 & 62.80 & 42.41 & 37.68 & 63.73 & 50.00 & 61.81 & 49.12 & 64.01 & 36.26 & 47.13 & 32.34 & 42.94 & 44.50 & 39.67 \\\hline
\textbf{BridG MT w/ Qwen2-72b} &  &  &  &  &  &  &  &  &  &  &  &  &  &  &  &  &  &  \\
\makecell[r]{Post Filtering} & 67.35 & 39.76 & 51.40 & 35.85 & 62.65 & 44.52 & 40.70 & 65.72 & 55.27 & 61.33 & 49.45 & 63.48 & 32.96 & 45.63 & 31.26 & 43.71 & 44.01 & 39.52 \\
\makecell[r]{Pre \& Post Filtering} & 67.11 & 39.51 & 51.31 & 35.48 & 62.56 & 44.42 & 40.48 & 65.76 & 55.28 & 61.52 & 49.30 & 63.48 & 32.96 & 45.61 & 31.26 & 43.68 & 43.96 & 39.04 \\
\textbf{BridG MT w/ Llama-3.1-8b} &  &  &  &  &  &  &  &  &  &  &  &  &  &  &  &  &  &  \\
\makecell[r]{Post Filtering} & 67.24 & 39.69 & 51.53 & 35.61 & 62.62 & 44.31 & 40.34 & 66.16 & 55.26 & 61.63 & 49.39 & 63.49 & 32.90 & 45.24 & 30.70 & 42.99 & 43.52 & 38.10 \\
\makecell[r]{Pre \& Post Filtering} & 67.20 & 39.31 & 51.44 & 35.04 & 62.56 & 44.24 & 40.18 & 66.01 & 55.32 & 61.61 & 48.99 & 63.81 & 32.90 & 45.05 & 30.97 & 43.05 & 43.52 & 37.87 \\
\textbf{BridG MT w/ Llama-3.2-3b} &  &  &  &  &  &  &  &  &  &  &  &  &  &  &  &  &  &  \\
\makecell[r]{Post Filtering} & 67.18 & 39.64 & 51.32 & 35.37 & 62.50 & 44.20 & 40.83 & 66.06 & 55.44 & 61.54 & 49.33 & 63.65 & 32.83 & 45.05 & 30.33 & 42.87 & 43.05 & 38.10 \\
\makecell[r]{Pre \& Post Filtering} & 67.17 & 39.47 & 51.33 & 35.09 & 62.44 & 44.13 & 40.82 & 65.92 & 55.57 & 61.45 & 49.34 & 63.65 & 32.83 & 44.94 & 30.51 & 42.88 & 43.04 & 38.06 \\
\hline
\textbf{QA Decode}&67.26	&39.65	&51.19	&35.31&	62.68&	44.28	&41.06&	66.47	&56.82	&62.92	&50.60&	63.47	&33.81&	41.99	&27.75&	41.76&	38.07	&34.05\\
\textbf{MAPS} &67.29	&39.69&	-&	-	&-	&-	&-&	65.17	&-&	47.66&	-&	54.36	&34.68&	-&	-	&-&	-&	-\\\hline
\end{tabular}}
\end{center}

\end{table*}

\begin{figure*}[ht]
    \centering
    \includegraphics[width=1\linewidth]{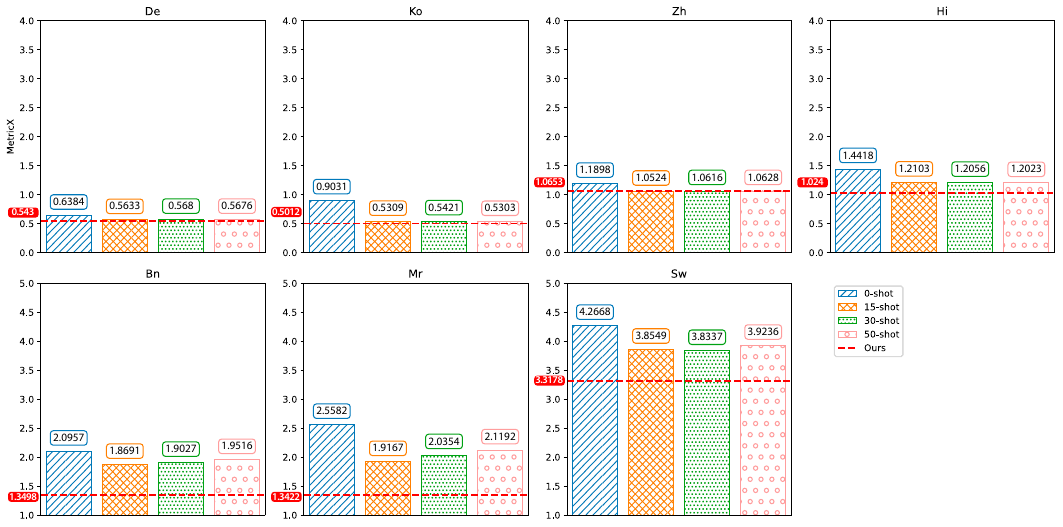}
    \caption{MetricX scores for few-shot translations (bars) and BridG MT translations (horizontal line). Lower score indicates better performance. Both BridG MT and few-shot scores represent the average MetricX scores of GPT-3.5 and Mistral-Nemo.}
    \label{fig:fewshot_compare_metricx}
\end{figure*}

\section{Results from Ablation}
\label{sec:ablation_full_result}
Table \ref{tab:ablation_result} shows the results for every combination of strategies that we explored in the ablation study(\ref{subsec:ablation_study}) with En-Ko translation. Table \ref{tab:adoption_all} presents the average results and changes in xCOMET scores for each combination of start selection strategies, the number of start sentences, and MT aggregation strategies.

\begin{table*}[ht]
\caption{Full results on ablation study with En-Ko translation task. All scores are measured with DA score by xCOMET. `Sort', `Filter', and `Tops' denote the start sentence selection methods. Each letter in parentheses denotes a similarity metric, with the order indicating the priority of the metrics. `L' stands for Levenshtein distance, `T' stands for tree edit distance, and `S' stands for SBERT similarity. The highest results for each start selection strategy are highlighted in bold, while the second-highest results are underlined.}
\label{tab:ablation_result}
\begin{center}
\resizebox{0.55\textwidth}{!}{
\begin{tabular}{cccccc}  
\multirow{2}{*}{\centering \makecell{\textbf{Start}\\\textbf{Selection}}} & \multirow{2}{*}{\centering \textbf{Filtering}} & \multicolumn{3}{c}{\textbf{Aggregation}}&\multirow{2}{*}{\centering \makecell{\textbf{Baseline}\\\textbf{(3shot)}}}  \\\cline{3-5}
& &\textbf{None} & \textbf{Poll} & \textbf{Prompt} &  \\\hline
\multirow{4}{*}{\centering Sort (S-T)} & All & 90.69 & 90.42 & 91.22 & \multirow{4}{*}{\centering 90.26} \\
 & Pre & 91.31 & 91.14 & 91.50 &\\
 & Post & \underline{92.29} & 92.18 & \textbf{92.54} &\\
 & Pre\&Post & 91.70 & 91.84 & 92.08 &\\\hline
\multirow{4}{*}{\centering Sort (T-S)} & All & 89.93 & 90.02 & 90.93 & \multirow{4}{*}{\centering 90.57} \\
 & Pre & 90.31 & 90.71 & 91.24 &\\
 & Post & 91.81 & 91.97 & \textbf{92.47} &\\
 & Pre\&Post & 91.74 & 91.64 & \underline{91.98} &\\\hline
\multirow{4}{*}{\centering Sort (L-S)} & All & 90.38 & 90.27 & 90.86 & \multirow{4}{*}{\centering 90.27} \\
 & Pre & 91.10 & 90.95 & 91.12 &\\
 & Post & \underline{92.08} & 91.96 & \textbf{92.41} &\\
 & Pre\&Post & 91.83 & 91.68 & 91.91 &\\\hline
\multirow{4}{*}{\centering Sort (T-L-S)} & All & 90.17 & 89.99 & 91.32 & \multirow{4}{*}{\centering 90.44} \\
 & Pre & 90.54 & 90.29 & 91.53 &\\
 & Post & 92.06 & 91.90 & \textbf{92.69} &\\
 & Pre\&Post & 92.00 & 91.81 & \underline{92.20} &\\\hline
\multirow{4}{*}{\centering Sort (L-T-S)} & All & 90.13 & 90.11 & 90.77 & \multirow{4}{*}{\centering 90.38} \\
 & Pre & 90.69 & 90.85 & 91.13 &\\
 & Post & 91.99 & \underline{92.05} & \textbf{92.24} &\\
 & Pre\&Post & 91.93 & 91.82 & 91.86 &\\\hline
\multirow{4}{*}{\centering \makecell{SBERT Filter\\+ Sort (T-L)}} & All & 90.36 & 90.23 & 91.07 & \multirow{4}{*}{\centering 90.35} \\
 & Pre & 91.04 & 90.88 & 91.30 &\\
 & Post & 92.07 & \underline{92.10} & \textbf{92.53} &\\
 & Pre\&Post & 91.70 & 91.69 & 92.00 &\\\hline
\multirow{4}{*}{\centering \makecell{SBERT Filter\\+ Sort (L-T)}} & All & 90.92 & 90.55 & 91.19 & \multirow{4}{*}{\centering 90.06} \\
 & Pre & 91.46 & 90.98 & 91.23 &\\
 & Post & \underline{92.37} & 92.29 & \textbf{92.61} &\\
 & Pre\&Post & 91.95 & 91.12 & 91.34 &\\\hline
\multirow{4}{*}{\centering Tops} & All & - & 90.12 & 90.98 & \multirow{4}{*}{\centering 90.10} \\
 & Pre & - & 90.65 & 91.30 &\\
 & Post & - & \underline{92.10} & \textbf{92.46} &\\
 & Pre\&Post & - & 92.06 & 92.01 &\\\hline
\end{tabular}}
\end{center}
\end{table*}

\begin{table*}[ht]
\caption{Average xCOMET scores and score changes of selected outputs when applying \textit{every combination} of strategies for start selection and aggregation in En-Ko translation. `Score change' is calculated only for the adopted outputs. `All' selects every output. `Pre-filtering', denoted as `Pre', applies zero-shot translation first and uses BridG MT only when the CometKiwi score is below a threshold. `Post-filtering', denoted ans `Post', selects outputs only if they outperform zero-shot translations. `Pre- \& Post- filtering', denoted as `Pre \& Post' combines these two strategies. We also report the number of end sentences for which bridging and Gradual MT is executed (`No. of Interpolated End Sents') and the number of end sentences for which the BridG MT output is selected over zero-shot translation (`No. of Selected Outputs'). The results show that `Pre \& Post' reduces the number of bridging by more than half while maintaining nearly the same translation performance.}
\label{tab:adoption_all}
\begin{center}
\resizebox{0.7\linewidth}{!}{\begin{tabular}{rrrrr}  
\makecell[l]{\textbf{Filtering}\\\textbf{Strategy}} &\makecell[l]{\textbf{Avg. Score}} & \makecell[l]{\textbf{Avg. Score Change} \\\textbf{of Selected Outputs}}& \makecell[l]{\textbf{No. of}\\\textbf{Interpolated}\\\textbf{End Sents(\%)}} & \makecell[l]{\textbf{No. of} \\\textbf{Selected Outputs} \\\textbf{(\%)}}\\\hline
Zeroshot & 89.48 & - & - & -\\
All & 90.55 & 1.06 & \makecell[r]{21864 (100\%)} & \makecell[r]{21864 (100\%)}\\
Pre & 91.01 & 3.70 & \makecell[r]{10468 (48\%)} & \makecell[r]{10468 (48\%)}\\
Post & \textbf{92.21} & 5.30 & \makecell[r]{21864 (100\%)} & \makecell[r]{11270 (52\%)}\\
Pre\&Post & 91.81 & \textbf{7.98} & \makecell[r]{10468 (48\%)} & \makecell[r]{6664 (30\%)}\\\hline
\end{tabular}}
\end{center}
\end{table*}

\clearpage

\section{Evaluation Results of BridG MT with Varying Start Pool Quality}
\label{sec:startpool_qual_result}
\begin{figure*}[ht]
    \centering
    \includegraphics[width=0.9\linewidth]{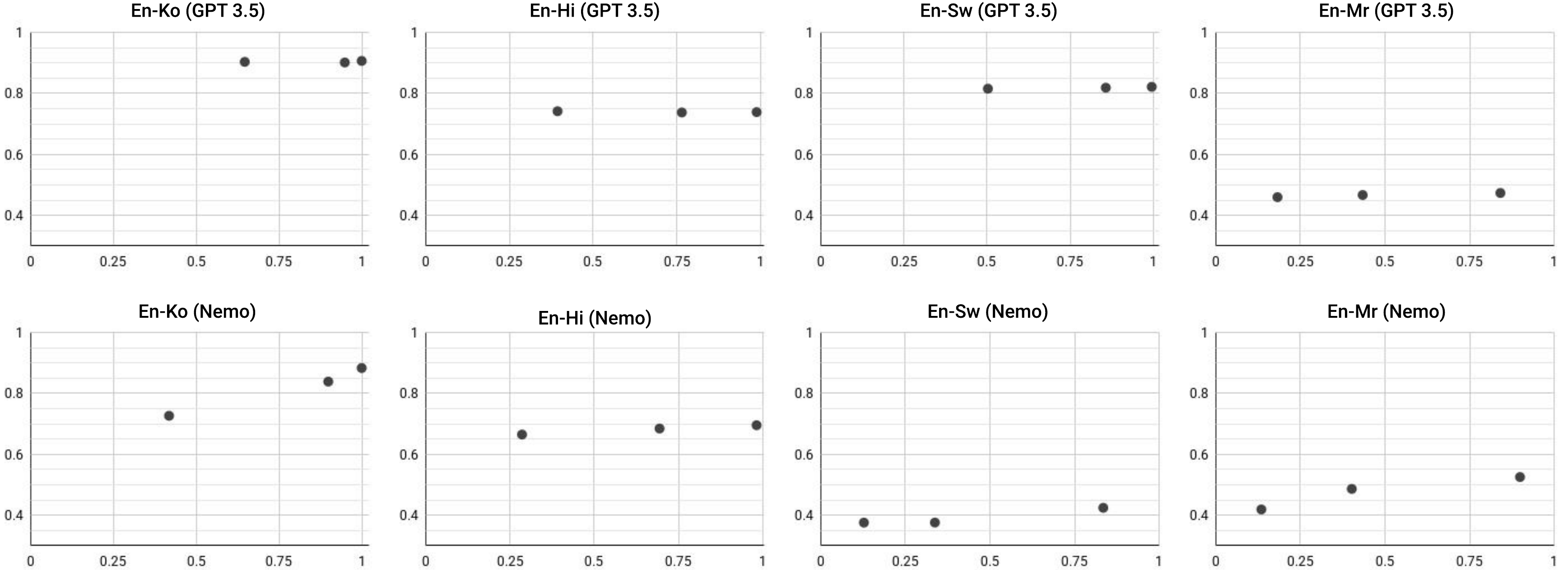}
    \caption{xCOMET scores of BridG MT when using start pools of varying quality. In each plot, the horizontal axis represents the xCOMET score of the start pool, while the vertical axis represents the xCOMET score of the final output. A single start sentence was used for evaluation, without the application of any filtering method.}
    \label{fig:startpool_qual}
\end{figure*}

 Figure \ref{fig:startpool_qual} shows xCOMET scores of BridG MT when using start pools of varying quality.

\section{Analysis On Sentence Bridging}
\label{sec:analysis_sentence_bridging}
Below we present the analysis we conducted on sentence bridges. We analyzed sentence bridges from En-Ko translation results.

\subsection{Progresses of Bridging.}
We examined whether LLMs genuinely bridge start and end sentences or simply generate random sentences. To assess this, we use SBERT to embed the sentences in bridge and calculate their Euclidean distances from the end sentence. If these distances generally decrease, it indicates successful bridging. To mesure this, we defined \textit{progress} as $\text{progress} = d_{n-1,e} - d_{n,e}$ where $d_{n,e}$ denotes the euclidian distance between $n$\textsuperscript{th} sentence in each bridge and end sentence.

\begin{figure*}[ht]
    \centering
    \includegraphics[width=1\linewidth]{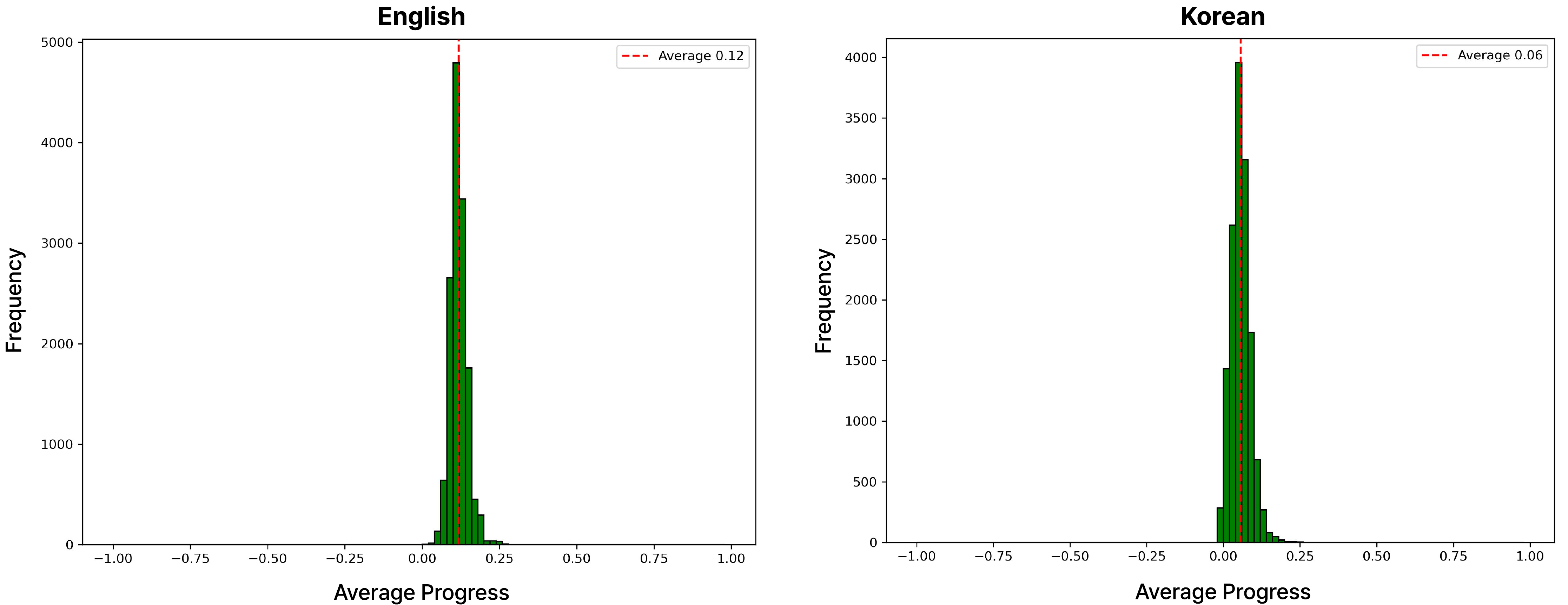}
    \caption{Distribution of average progress from each sentence bridge. The sentence bridges are drawn from the English-to-Korean BridG MT results. Progress indicates how much each sentence in the bridge moves closer to the target sentence in terms of Euclidean distance.}
    \label{fig:average_progress_enko}
\end{figure*}

As shown in Figure \ref{fig:average_progress_enko}, the average progress of each bridge is generally positive on both the source(English) and target(Korean) sides, indicating that the sentence bridging effectively connects the two sentences. The mean and standard deviation of average progress across bridges were 0.12 and 0.27 for English, and 0.06 and 0.031 for Korean, respectively.

\subsection{Visualization of Bridges}
\label{subsec:bridges_visualization}

\begin{figure*}[ht]
    \centering
    \includegraphics[width=1\linewidth]{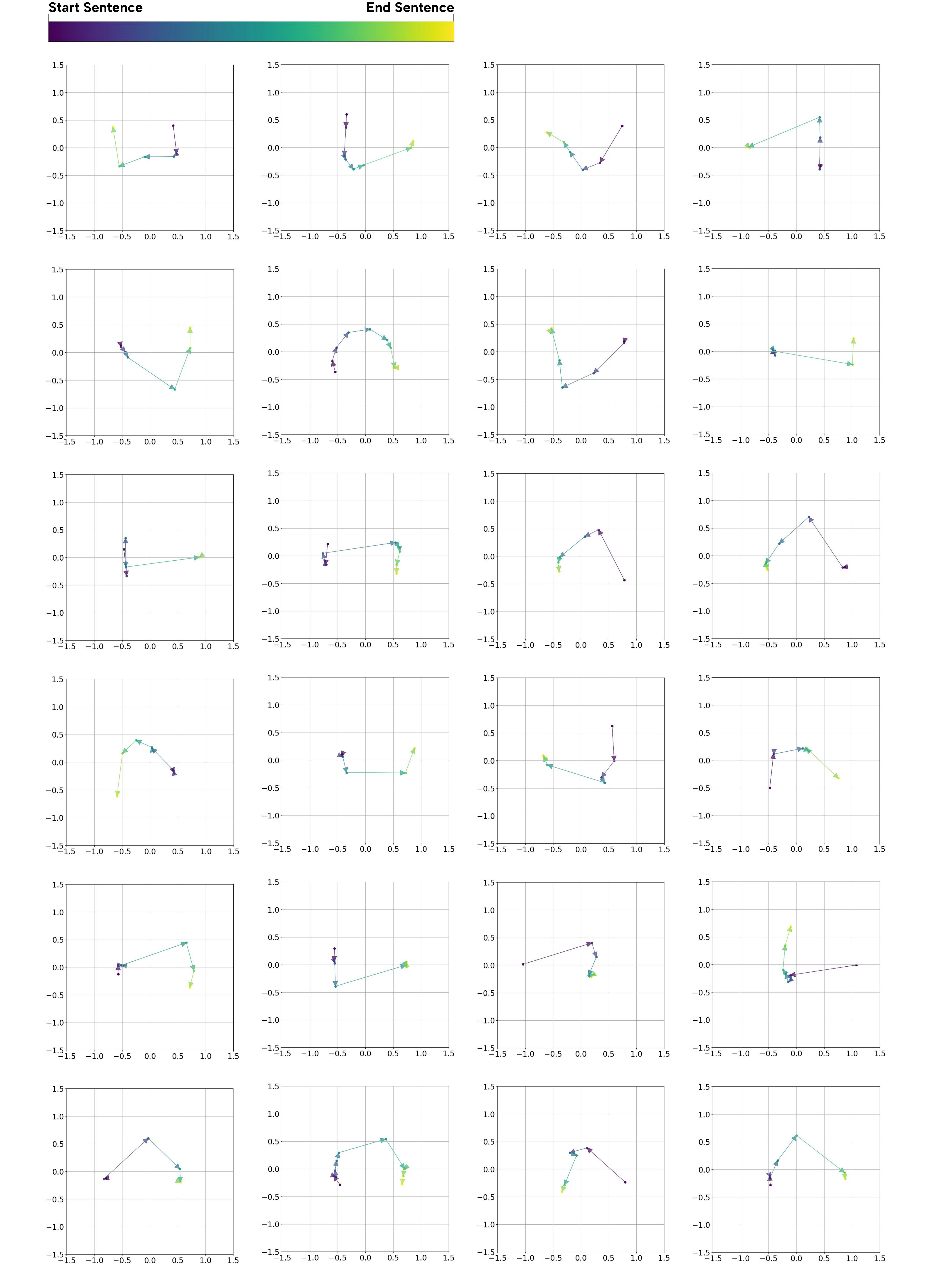}
    \caption{24 samples of 2D scatter plots of embeddings from sentence bridges. The plots are projected from SBERT embeddings onto a 2D plane using PCA. The X and Y axes of each plot represent the first and second principal components, respectively. Arrows in each plot show the trajectory of sentence shifts from the start sentence (blue-colored dot) to the end sentence (yellow-colored dot).}
    \label{fig:pca_patterns}
\end{figure*}

\begin{figure*}[ht]
    \centering
    \includegraphics[width=1\linewidth]{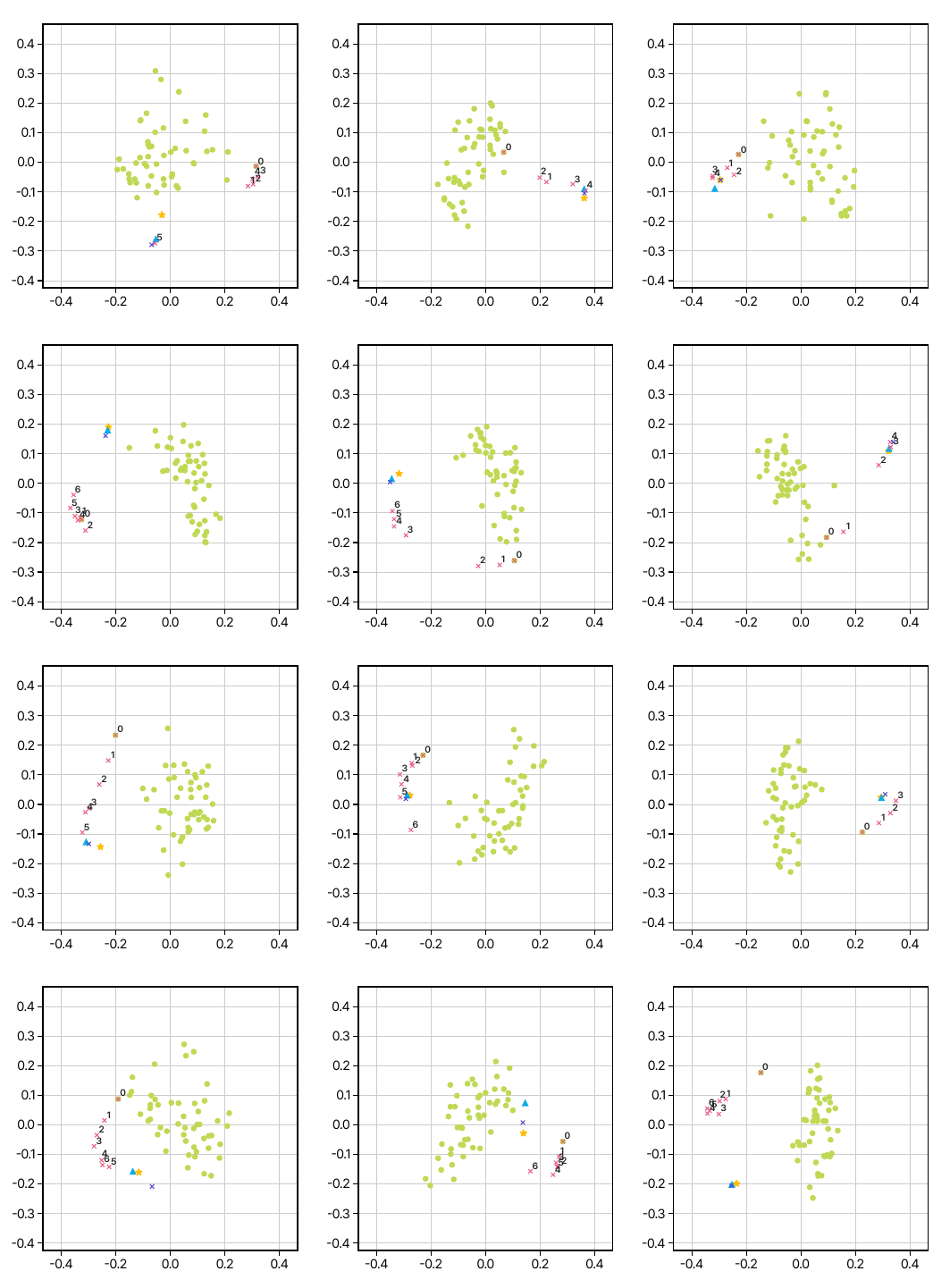}
    \caption{12 2D scatter plots of embeddings from Gradual MT. Sentence embeddings from the intfloat/multilingual-e5-large model are projected onto a 2D plane using PCA. The X and Y axes represent the first and second principal components, respectively. Green dots and sky-blue triangles indicate the few-shot examples and 50-shot MT outputs. Purple and blue X marks denote the translations from each step and the final translation generated by Gradual MT.}
    \label{fig:pca_patterns_tgt}
\end{figure*}

In addition to the analysis on progress, we conducted visualizations on both the source and target sides to examine whether our experimental results align with the intuition illustrated in Figure 2. Specifically, we visualized the embeddings of the sentences within the sentence bridges and the Korean translations generated at each step of Gradual MT. For English sentences, we used all-mpnet-base-v2 to obtain embeddings, while for Korean sentences, we used intfloat/multilingual-e5 \citep{wang2024multilingual}. On the Korean side, we additionally plotted 50-shot example translations to examine whether the intermediate translations produced at each Gradual MT step gradually move closer to the embedding of the gold translation, and how they compare to the few-shot examples. The results of this analysis show that sentence bridging effectively generates intermediate sentences that bridge the start and end sentences on the source side. Furthermore, we frequently observed that the intermediate translations from Gradual MT also move progressively closer to the gold translation. These findings support the intuition underlying our approach. The visualizations are shown in Figure \ref{fig:pca_patterns} and Figure~\ref{fig:pca_patterns_tgt}.

\end{document}